\documentclass[runningheads]{llncs}

\usepackage{eccv}

\usepackage{eccvabbrv}

\usepackage{graphicx}
\usepackage{booktabs}
\usepackage{cite}
\usepackage{color}
\usepackage{caption}
\usepackage{float}
\usepackage{array}
\usepackage{makecell}
\usepackage{multirow}

\usepackage[accsupp]{axessibility}  
\usepackage{xcolor,colortbl}

\usepackage{hyperref}

\usepackage{orcidlink}

\usepackage{tikz}
\usetikzlibrary{positioning,fit,backgrounds,calc}

\begin{document}

\title{Comprehension of Multilingual Expressions Referring to Target Objects in Visual Inputs} 

\titlerunning{Multilingual Referring Expression Comprehension}

\author{Francisco Reis Nogueira\inst{1} \and
Bruno Martins\inst{1} \and
Alexandre Bernardino\inst{2}}

\authorrunning{F.~R.~Nogueira et al.}

\institute{INESC-ID, Instituto Superior T\'ecnico, Universidade de Lisboa, Portugal \and
ISR-Lisboa, LARSyS, Instituto Superior T\'ecnico, Universidade de Lisboa, Portugal\\
{\footnotesize\email{\{francisco.r.nogueira,bruno.g.martins,alexandre.bernardino\}@tecnico.ulisboa.pt}}}

\maketitle

\begin{abstract}
Referring Expression Comprehension (REC) requires models to localize objects in images based on different types of natural language descriptions. Even with significant progress, research on the area remains predominantly English-centric, despite increasing global deployment demands. This work addresses multilingual REC through two main contributions. First, we construct a unified multilingual dataset spanning 10 languages, by systematically expanding 12 existing English REC benchmarks through machine translation and context-based translation enhancement. Second, we introduce an attention-anchored efficient neural architecture that uses a multilingual SigLIP2 encoder. Our attention-based approach generates coarse spatial anchors from attention distributions, which are subsequently refined through learned residuals. Experimental evaluation demonstrates competitive performance on standard benchmarks, despite the use of a relatively small model. Multilingual evaluation shows consistent capabilities across languages, establishing the practical feasibility of efficient multilingual visual grounding systems.
\end{abstract}

\vspace{-0.25cm}

\section{Introduction}
\label{sec:intro}

\vspace{-0.25cm}

Humans can naturally identify objects in visual scenes from context dependent referring expressions. A simple phrase like "\textit{the red sports car}" or a complex description like "\textit{the parked ketchup rocket}" can both reference the same object within a visual scene. This capability extends across cultural and linguistic boundaries, with users adjusting expressions to accommodate common sense. In the context of vision and language research, this capability is studied through Referring Expression Comprehension (REC) tasks.

Addressing REC requires robust cross-modal understanding and spatial language grounding \cite{tumu2025referringexpressionslensspatial}, through AI systems that must simultaneously parse linguistic information and accurately map it to visual cues within images. In this context, the development of encoder models based on Contrastive Language-Image Pre-training (CLIP) substantially improved text-image alignment, enabling progress in different vision and language tasks \cite{radford2021learning}. Multilingual CLIP (mCLIP) further extended these capabilities, enabling richer connections between multilingual text encoding and visual representations \cite{chen_mclip_2023}. Recent state-of-the-art vision and language encoder models, such as SigLIP2 \cite{tschannen2025siglip2multilingualvisionlanguage}, are also natively multilingual, and they currently support many multimodal LLMs.

Despite the aforementioned advances, current REC research faces a critical limitation: nearly all benchmark datasets and models operate exclusively in English, which creates several fundamental limitations. First, referring expressions are inherently tied to linguistic structure, while spatial prepositions, adjective ordering, and descriptive conventions can vary significantly across languages. This means that models trained on English data may fail to generalize to other linguistic patterns. Second, simple machine translation introduces semantic shifts that can compromise grounding accuracy, limiting translation plus grounding pipelines. Phrases that convey clear meaning in one language may become ambiguous when translated to English, or carry culturally distinct connotations that alter the visual interpretation. Standard REC benchmarks such as RefCOCO \cite{yu2016modeling} or RefCLEF \cite{kazemzadeh-etal-2014-referitgame}, while providing essential evaluation frameworks, remain overwhelmingly monolingual. This limitation restricts research on cross-lingual grounding capabilities and prevents the development of vision-language systems that can serve the global population in native languages.

Our work addresses the aforementioned limitations through two main contributions: (1) a new multilingual dataset built by expanding 12 established English REC datasets into a generalized and unified multilingual corpus spanning 10 languages, and (2) an efficient and multilingual attention-anchored neural architecture to guide bounding box generation. We trained a model following the proposed architecture and with the unified dataset, and show that it achieves competitive performance on standard benchmarks, while ensuring consistent multilingual capabilities. For instance, despite the small model size, on RefCOCO \cite{yu2016modeling} we achieve an averaged result of 91.3\% IoU@50 in English evaluation, with a limited degradation across the other languages. The model is particularly consistent among Romance languages (i.e., Spanish, French, Italian, and Portuguese), with mean IoU differences of only 2-4\% compared to English. These results validate the feasibility of efficient multilingual REC systems that can serve users in their native languages.

The remainder of this paper is organized as follows: Section 2 reviews related work. Section 3 details our contributions, including the dataset construction through machine translation and content-aware quality enhancement (Section 3.1), and the attention-anchored neural network architecture (Section 3.2). Section 4 presents the experimental evaluation. Finally, Section 5 summarizes our findings and outlines directions for future work.
\vspace{-0.25cm}

\section{Related Work}
\label{sec: backg}

\vspace{-0.25cm}

Referring Expression Comprehension (REC) requires joint reasoning over the visual and linguistic modalities. This section reviews foundational developments in vision-language alignment, object detection architectures, and previous REC methods that inform our approach.

\vspace{-0.25cm}

\subsection{Vision-Language Foundation Models}

\vspace{-0.25cm}

Contrastive Language-Image Pre-training (CLIP) demonstrated that contrastive learning on large-scale image-text pairs produces powerful joint representations in a shared embedding space, enabling zero-shot transfer \cite{radford2021learning}. Multilingual extensions followed. For instance, mCLIP achieved cross-lingual vision-language alignment through triangular knowledge distillation from English-trained encoders to multilingual text encoders \cite{chen_mclip_2023}. SigLIP refined the training approach by replacing the contrastive loss with a sigmoid loss, treating image-text matching as independent binary classification \cite{zhai2023sigmoidlosslanguageimage}. Building on this foundation, SigLIP2 incorporated decoder-based pretraining for localization tasks, different self-supervised losses, and multilingual data curation on the WebLI dataset spanning 100+ languages \cite{tschannen2025siglip2multilingualvisionlanguage}. The SigLIP2 multilingual vision-language encoder provides the foundation for our attention-anchored architecture.

\vspace{-0.35cm}

\subsection{Object Detection and REC Architectures.}

\vspace{-0.25cm}

Previous REC methods build upon general object detection frameworks, adapting localization mechanisms for language-guided tasks. While traditional approaches like Faster R-CNN employed region proposal networks with predefined anchors \cite{ren2016fasterrcnnrealtimeobject}, DEtection TRansformers (DETR) introduced learned object queries for end-to-end detection \cite{carion_end--end_2020}. Subsequent work showed that anchor-based approaches remain effective when integrated with Transformers. For instance, DAB-DETR used dynamic anchor boxes as explicit positional queries that refine iteratively \cite{liu2022dabdetrdynamicanchorboxes}, while AnchorDETR demonstrated that learning to generate anchor points from queries enhances detection, particularly for small objects \cite{wang2022anchordetrquerydesign}. These findings established that anchors derived from learned representations, rather than fixed heuristics, can provide valuable inductive biases.

Early REC methods used combinations of convolutional and recurrent networks but struggled with complex linguistic structures \cite{mao2016generationcomprehensionunambiguousobject}. TransVG marked the transition to end-to-end Transformer models for REC, reformulating visual grounding as direct coordinate regression, and demonstrating that simple Transformer stacks, with self-attention over concatenated features, outperform elaborate fusion mechanisms \cite{deng2021transvg}. ReferFormer introduced the "\textit{language as queries}" paradigm where linguistic expressions directly condition object queries, forcing focus on the referred object \cite{wu2022languagequeries}. Recent architectures prioritize efficiency and open-vocabulary capabilities, for instance with Grounding DINO extending detection to open-set scenarios \cite{liu2023groundingdinomarryingdino}, although remaining limited to English descriptions. Moreover, recent combinations of Large Language Models (LLMs) with vision encoders like CLIP/SigLIP support grounding tasks through text instructions, although the huge size of these models can limit applications involving hardware constraints, or requiring the efficient analysis of datasets.

Despite the many recent advances, multilingual visual grounding research remains very limited. Referring expressions exhibit language-specific patterns in prepositions, adjective ordering, and descriptive conventions \cite{waldon-degen-2021-modeling}, while standard benchmarks like RefCOCO or RefCLEF remain overwhelmingly English-centric \cite{lin2015microsoftcococommonobjects, kazemzadeh-etal-2014-referitgame}. Our architecture builds on attention-based spatial reasoning from Transformer detection, while adapting these methods for multilingual deployment through frozen multilingual encoders.

\vspace{-0.3cm}

\section{The Proposed Approach to Generalized Multilingual Referring Expression Comprehension}
\label{sec:methodology}

\vspace{-0.3cm}

This section presents our approach to multilingual REC. We first describe the construction of a unified multilingual dataset that systematically expands existing English REC benchmarks across 10 languages, through automatic translation, validation, and quality enhancement. We then introduce an attention-anchored neural architecture that uses multilingual vision-language encoders, together with efficient cross-modal enhancement and spatial attention mechanisms, for robust cross-lingual localization.

\begin{table}[!t]
\vspace*{-0.4cm}
\centering
\caption{Overview on the composition of the unified multilingual dataset, showing statistics for the 12 integrated benchmarks that, together, yield approximately 8M referring expressions across 177,620 images and 10 languages.}
\label{tab:datasets}

\renewcommand{\arraystretch}{1.2}
\setlength{\tabcolsep}{3pt}
\resizebox{\textwidth}{!}{%

\begin{tabular}{l|c|ccc|ccccc}
\toprule
\makecell{ \\\textbf{Dataset}} &
\makecell{\textbf{Number of}\\\textbf{Instances}} & 
\makecell{ \\\textbf{Train}} &
\makecell{ \\\textbf{Validation}} &
\makecell{ \\\textbf{Test}} &
\makecell{ \\\textbf{Images}} &
\makecell{ \\\textbf{Objects}} &
\makecell{ \\\textbf{Expressions}} & 
\makecell{\textbf{English}\\\textbf{Expressions}} & \makecell{\textbf{Image}\\\textbf{Source}} \\
\midrule
RefOI & 4,920 & 3,940 & 490 & 490 & 488 & 492 & 4,920 & 492 & OpenImages \\
Multimodal Ground & 8,550 & 7,040 & 740 & 770 & 2,550 & 855 & 8,550 & 855 & Multimodal Ground \\
LocateBench & 13,170 & 10,530 & 1,320 & 1,320 & 1,317 & 1,317 & 13,170 & 1,317 & MS COCO \\
FG-OVD & 23,490 & 9,390 & 13,560 & 6,780 & 1,707 & 2,349 & 23,490 & 2,349 & MS COCO \\
RefDrone & 170,330 & 123,360 & 14,210 & 32,760 & 7,971 & 16,938 & 170,330 & 17,033 & VisDrone \\
Hindi Visual Genome & 315,200 & 289,270 & 9,890 & 15,950 & 31,520 & 31,520 & 315,200 & 31,520 & Visual Genome \\
Toloka & 451,990 & 389,900 & 17,050 & 45,040 & 45,199 & 45,199 & 451,990 & 45,199 & MS COCO \\
RefCOCOg & 950,100 & 740,170 & 209,930 & 104,965 & 25,799 & 49,820 & 950,100 & 95,010 & MS COCO \\
RefCLEF & 1,303,640 & 1,140,980 & 25,030 & 37,630 & 19,997 & 29,985 & 1,303,640 & 130,364 & SAIAPR-12 \\
RefCOCO+ & 1,415,640 & 1,201,910 & 107,580 & 106,150 & 19,992 & 49,855 & 1,415,640 & 141,564 & MS COCO \\
RefCOCO & 1,422,100 & 1,384,470 & 10,340 & 27,290 & 19,944 & 49,999 & 1,422,100 & 142,210 & MS COCO \\
FineCops-Ref & 1,918,520 & 1,637,920 & 184,550 & 96,050 & 39,675 & 68,241 & 1,918,520 & 191,852 & GQA \\
\midrule
\rowcolor{gray!15}
\textbf{Overall} & 7,997,650 & 6,938,880 & 594,690 & 475,195 & 177,620 & 336,882 & 7,997,650 & 799,765 & --- \\
\bottomrule
\end{tabular}
}
\end{table}

\vspace{-0.3cm}

\subsection{Dataset Collection and Integration}

\vspace{-0.3cm}

Our dataset construction methodology follows three guiding principles: (1) visual diversity, spanning everyday scenes, aerial/overhead imagery, fine-grained attributes, and compositional reasoning; (2) expression variety, ranging from concise phrases to elaborate descriptions and question-based formats; and (3) linguistic breadth, extending all English content to 9 additional languages, using automatic quality estimation and targeted re-translation to improve the preservation of referential accuracy.

We integrated 12 publicly available English datasets totaling 799,765 unique expressions across 177,620 images. This collection encompasses multiple visual domains and expression styles: the RefCOCO family provides standard everyday scene benchmarks with varying constraint sets \cite{yu2016modeling}; RefCLEF adds natural stylistic variation through gameplay collection \cite{kazemzadeh-etal-2014-referitgame}; FineCops-Ref and FG-OVD test fine-grained attribute discrimination \cite{liu2024finecopsrefnewdatasettask, bianchi2024devil}; RefDrone introduces aerial perspectives with small objects \cite{sun2025refdronechallengingbenchmarkreferring}; Hindi Visual Genome emphasizes ambiguous and underspecified references requiring contextual disambiguation \cite{hindi-visual-genome:2019}; RefOI isolates basic grounding without distractors, using single-presence instances \cite{refoi-ma2025vision}; Multimodal Ground targets instance-level product identification in controlled retail environments \cite{lu2025multimodalreferencevisualgrounding}; finally, Toloka VQA and LocateBench explore interrogative and multiple-choice formats \cite{TolokaWSDMCup2023, chiang2024locatebenchevaluatinglocatingability}.

The integration provides 799,765 unique English expressions which we translated to 9 additional languages (i.e., Chinese, German, Dutch, Spanish, French, Italian, Korean, Portuguese, and Russian), yielding approximately 8 million total multilingual expressions across 177,620 images. The selected 9 languages cover approximately 3.5 billion native speakers, spanning five major families (i.e., Romance, Germanic, Slavic, East Asian, and Koreanic) and aligning with the languages well-supported by the models used in our work (i.e., the models for machine translation and quality estimation, and also SigLIP2).

All datasets were standardized to a common JSON format, preserving bounding boxes in original pixel coordinates, image metadata, and dataset provenance. For datasets with existing train/validation/test splits, we mostly maintained the original partitions to ensure comparability with published baselines. However, in order to avoid contamination problems, we removed from the training split all the images that also occurred in the testing splits of other datasets. For LocateBench, which lacked predefined splits, we applied deterministic shuffling with a fixed random seed to create reproducible 80/10/10 train/validation/test divisions. In the case of FG-OVD and the Google variant of RefCOCOg, the official test partitions were unavailable in the original source distributions. To enable comprehensive evaluation while maintaining methodological rigor, we strategically allocated half of each validation set to serve as the corresponding test partition for these datasets. For RefCOCO and RefCOCO+, our test split aggregates the original testA and testB dataset splits.

After initial data collection we proceed through three stages: (1) translation of all English expressions to 9 languages using specialized multilingual models; (2) use of reference-free translation quality estimation to identify low-quality translations; (3) perform a multimodal enhancement that incorporates visual context for targeted improvement.

\vspace{-0.4cm}

\subsubsection{Machine Translation Methodology:}

We employed TowerInstruct-7B-v0.2, i.e. a multilingual language model which specializes on machine translation, as our primary translation engine \cite{tower_llm_2024}. The TowerInstruct-7B model handles multiple translation-related tasks, including sentence-level translation, terminology-aware adaptation, and context-aware phrasing, making it well suited for referring expression translation. The model is also computationally efficient, compared to the use of much larger frontrier language models. We translated each English expression, in each source dataset, to all the considered 9 target languages (i.e., Chinese, German, Dutch, Spanish, French, Italian, Korean, Portuguese, and Russian), using a consistent prompt structure: "Translate to \{target\_language\}.\textbackslash n English: \{text\}\textbackslash n \{target\_language\}:". The complete translation process generated approximately 7.2 million translation pairs, producing a dataset of $\approx$ 8M total multilingual referring expressions, including the English sources.

\vspace{-0.4cm}

\subsubsection{Quality Validation and Enhancement:}

We employed COMETkiwi-DA for reference-free translation quality evaluation, accessing and scoring the 7.2M translation pairs without any manual annotation \cite{rei-etal-2022-cometkiwi}. Each language had a quality threshold defined as the 40th percentile ($P_{40}$), with score distributions strongly right-skewed toward the 0.75-0.85 range. Translations scoring below the language-specific $P_{40}$ threshold were flagged for targeted enhancement. Despite generally high quality scores, errors included over-literal translations losing the use of idiomatic expressions, translation failures generating contextually inappropriate responses, and semantic information loss omitting critical visual descriptors. These errors motivated the targeted enhancement strategy, in which we re-translated the $\approx$3.7M expressions scoring below $P_{40}$ (i.e., the lowest-quality 40\% instances per language) using Google's Gemini 2.0 Flash Lite with visual context included \cite{google_gemini2_flashlite_2025}. The enhancement pipeline processed each translation candidate using (1) the original English expression; (2) the initial text-only translation; and (3) the source image with the target object bounding box overlaid. Re-evaluation with COMETkiwi-DA demonstrated measurable improvements upon the enhancement across all languages. Threshold gains ranged from +0.01 (in Portuguese, Russian, and Spanish) to +0.02 (in Chinese), with enhanced subset medians reaching 0.80-0.83. The enhanced mean scores clustered at 0.81-0.84, significantly above the original thresholds. The use of the $P_{40}$ threshold follows from a preliminary manual inspection of 100 translations for some of the considered languages, sampled across the full range of quality scores, assessing whether each translation preserved the referential meaning, remained fluent in the target language, and retained the visual descriptors needed to identify the target object. Below approximately the 40th percentile these problems became frequent enough to justify the additional cost of re-translation, while higher-scoring translations were generally adequate for the task. We note that this inspection was carried out by the authors, and covers only the languages in which the authors are competent (i.e., Portuguese, Spanish, French, and Italian), constituting a coarse calibration. Our quality control is therefore mostly automatic, and a native-speaker evaluation of referential preservation across the ten languages remains an important direction for future work.

\vspace{-0.35cm}

\subsubsection{Characterization of the Resulting Dataset:}

The final unified dataset, summarized in Table \ref{tab:datasets}, comprises approximately 8 million multilingual referring expressions across 10 languages, derived from approximately 800K unique English expressions over 177,620 distinct images with 336,882 annotated objects. The dataset features substantial diversity in visual domains (everyday scenes: 65\%, aerial/overhead imagery: 2.1\%, fine-grained attributes: 0.3\%, instance-level objects: 4.1\%) and expression complexity levels (simple single-object references: 15-20\%, moderate 2-3 attributes/relations: 50-55\%, complex compositional expressions: 25-30\%). These statistics are derived from the original dataset characterizations, as given in the respective publications and content distributions.

\vspace{-0.3cm}

\subsection{A New Architecture for Referring Expression Comprehension}

\vspace{-0.25cm}

We propose a new attention-anchored neural architecture for Referring Expression Comprehension (REC), which decomposes the localization task into coarse spatial attention, and learned fine-grained refinement. The model combines frozen SigLIP2 encoders with trainable cross-modal fusion modules and spatial attention layers, producing normalized bounding box coordinates directly from text-image pairs, while still relying on a small number of parameters.

\begin{figure*}[!t]
    \centering
    \resizebox{\linewidth}{!}{
%
%
%
\begin{tikzpicture}[
  font=\sffamily\small,
  >=stealth,
  img/.style  ={rectangle, rounded corners=2pt, draw=magenta!40!black, fill=magenta!7,
                minimum height=10mm, minimum width=21mm, align=center, inner sep=2pt},
  txt/.style  ={rectangle, rounded corners=2pt, draw=orange!55!black, fill=orange!9,
                minimum height=10mm, minimum width=21mm, align=center, inner sep=2pt},
  attn/.style ={rectangle, rounded corners=2pt, draw=yellow!55!black, fill=yellow!25,
                minimum height=10mm, minimum width=21mm, align=center, inner sep=2pt},
  dec/.style  ={rectangle, rounded corners=2pt, draw=blue!60!black, fill=blue!12,
                minimum height=10mm, minimum width=21mm, align=center, inner sep=2pt},
  box/.style  ={rectangle, rounded corners=2pt, draw=violet!65!black, fill=violet!12,
                minimum height=10mm, minimum width=21mm, align=center, inner sep=2pt},
  gt/.style   ={rectangle, rounded corners=2pt, draw=green!45!black, fill=green!12,
                minimum height=10mm, minimum width=21mm, align=center, inner sep=2pt},
  loss/.style ={rectangle, rounded corners=2pt, draw=black, fill=black, text=white,
                minimum height=10mm, minimum width=44mm, align=center, inner sep=2pt},
  grp/.style  ={rectangle, rounded corners=3pt, draw=none, fill=black!8, inner sep=3mm},
  lbl/.style  ={font=\sffamily\scriptsize\bfseries, text=black!60, align=center},
  tag/.style  ={font=\sffamily\scriptsize, text=black!70},
  flow/.style ={->, thick, draw=black!58},
  sup/.style  ={->, thick, dashed, draw=black!45},
]

\node[img]                  (imgin)  at (0,0)      {Image};
\node[txt]                  (txtin)  at (0,-1.45)  {Referring\\expression};

\node[img]                  (venc)   at (3.05,0)     {Vision encoder\\\footnotesize 576 patches};
\node[txt]                  (tenc)   at (3.05,-1.45) {Text encoder\\\footnotesize text tokens};

\node[img]                  (fusev)  at (6.10,0)      {Image stream};
\node[txt]                  (fuset)  at (6.10,-1.45)  {Text stream};

\node[dec]                  (decod)  at (9.15,0)     {Cross-attention\\(two layers)};
\node[txt]                  (query)  at (9.15,-1.45) {Pooling + MLP\\\footnotesize single query};

\node[attn]                 (amap)   at (12.20,0)     {$24\times24$\\attention map};

\begin{scope}[on background layer]
  \node[grp, fit=(venc)(tenc), label={[lbl]above:{SigLIP2\\(mostly frozen)}}] {};
  \node[grp, fit=(fusev)(fuset), label={[lbl]above:{Bidirectional fusion\\(four layers)}}] {};
\end{scope}

\draw[flow] (imgin) -- (venc);
\draw[flow] (txtin) -- (tenc);
\draw[flow] (venc)  -- (fusev);
\draw[flow] (tenc)  -- (fuset);
\draw[<->, thick, draw=black!65] (fusev) -- (fuset) node[midway, right=1mm, tag] {cross-attn};
\draw[flow] (fusev) -- (decod) node[midway, above, tag] {K,\,V};
\draw[flow] (fuset) -- (query);
\draw[flow] (query) -- (decod) node[midway, right=1mm, tag] {Q};
\draw[flow] (decod) -- (amap);

\node[attn] (marg)   at (0,-4.35)    {Marginalize\\$p_x,\;p_y$};
\node[attn] (perc)   at (3.05,-4.35) {10th / 90th\\percentiles};
\node[box]  (anchor) at (6.10,-4.35)  {Anchor box};
\node[box]  (resid)  at (9.15,-4.35) {MLP residual\\\footnotesize logit space};
\node[box]  (pred)   at (12.20,-4.35) {Predicted\\box};

\begin{scope}[on background layer]
  \node[grp, fit=(marg)(perc)(anchor)(resid),
        label={[lbl]below:Attention-anchored grounding\ \ (anchor is non-differentiable)}] {};
\end{scope}

\draw[flow] (marg)   -- (perc);
\draw[flow] (perc)   -- (anchor);
\draw[flow] (anchor) -- (resid);
\draw[flow] (resid)  -- (pred);

\node[gt]   (gtbox) at (8.0,-6.55)  {Ground-truth\\box};
\node[loss] (loss)  at (12.20,-6.55)
      {Training loss\\[1pt]
       $\lambda_{L1}\mathcal{L}_{L1}+\lambda_{GIoU}\mathcal{L}_{GIoU}+\lambda_{attn}\mathcal{L}_{attn}$};

\draw[sup] (gtbox) -- (loss);

\draw[sup] (pred.south) -- (loss.north);
\node[tag, anchor=west] at (12.40,-5.45) {$\mathcal{L}_{L1},\ \mathcal{L}_{GIoU}$};

\draw[sup, rounded corners=5pt]
      (amap.east) -- (15.75,0) -- (15.75,-6.55) -- (loss.east);
\node[tag, anchor=south] at (14.45,0.20) {$\mathcal{L}_{attn}$ (KL)};

\draw[flow, rounded corners=5pt]
      (amap.south) -- ++(0,-2.15) -- ($(marg.north)+(0,0.90)$) -- (marg.north);

\end{tikzpicture}}
    \caption{Overview on the proposed attention-anchored architecture, featuring frozen SigLIP2 encoders, bidirectional cross-modal fusion, text query aggregation, a cross-attention decoder, and attention-guided bounding box prediction with residual refinement. The attention map is supervised through a KL divergence against the rasterized ground-truth box, while the predicted box is supervised both through coordinate regression and brounding box overlap losses.}
    \label{fig:ArchCompressed}
\end{figure*}
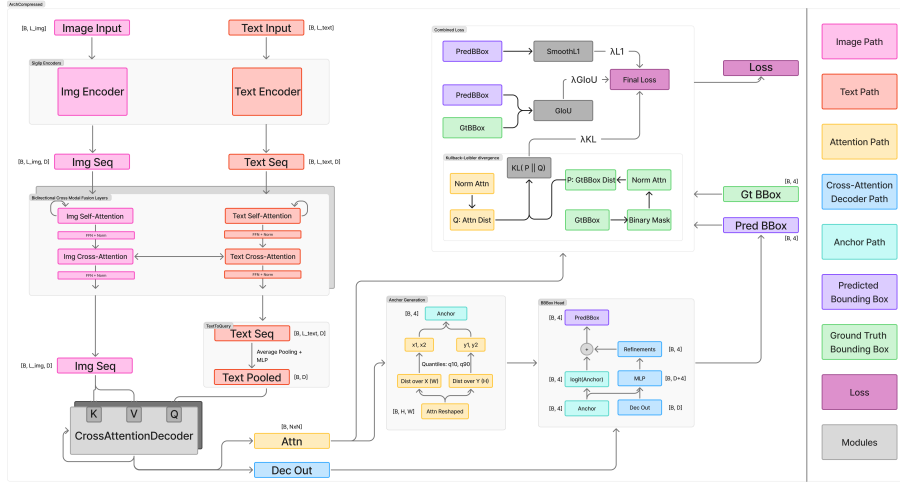

The architecture can be visualized in Figure \ref{fig:ArchCompressed}, which illustrates the overall path that an input would take within the model. Given an image and a natural language referring expression, the model predicts a bounding box in normalized coordinates that localizes the referred object. Unlike in the context of multi-object detection tasks, our REC formulation requires identifying exactly one target object specified by the linguistic description, often requiring fine-grained discrimination among visually similar candidates. The proposed architecture processes inputs through five stages, which will be defined next.

\vspace{-0.3cm}

\subsubsection{Feature Extraction with SigLIP2:}

We use SigLIP2 encoders\footnote{{\tiny\url{https://huggingface.co/google/siglip2-base-patch16-384}}} to extract aligned text and image representations. The text encoder processes tokenized sequences through transformer layers, producing contextualized representations for each token. The vision encoder divides 384x384 pixel images into $16 \times 16$ pixel patches, yielding a $24 \times 24$ grid of 576 visual tokens. While most parts of the base SigLIP2 encoders remain frozen to preserve pre-trained visual-linguistic alignment, we unfreeze the final three layers of both the text and vision encoders. This selective fine-tuning allows task-specific adaptation of the encoders, while maintaining general semantic understanding, which corresponds to a critical balance for multilingual generalization \cite{yosinski2014transferablefeaturesdeepneural}. Note that we have only used the base variant of SigLIP2, envisioning deployment in resource constrained settings. Experiments with larger encoder models are left for future work.

\vspace{-0.3cm}

\subsubsection{Bidirectional Cross-Modal Fusion:}

A four-layer bidirectional Transformer iteratively refines initial alignments between text and image representations through parallel processing streams. Each layer contains four distinct attention mechanisms, two of them refining the vision representations, and the other two refining the text representations: (1) text self-attention, where text tokens attend to other text tokens; (2) text-to-image cross-attention, where text tokens query image patches; (3) image self-attention, where image patches attend to other patches; (4) and image-to-text cross-attention, where image patches query text tokens. Each attention operation, within each layer, is followed by feed-forward networks with residual connections and layer normalization.

\vspace{-0.3cm}

\subsubsection{Text Query Aggregation: }

After refinement with the cross-attention layers, we consolidate the distributed text representation into a single compact query vector. This aggregation employs masked mean pooling over valid text tokens to create a global text representation, followed by one feed-forward projection layer to transform the text representation into a query space. A learnable bias vector, associated to the feed-forward layer, provides task-specific initialization, this way breaking symmetry in the query representation. Notice that unlike object detection methods like DETR, that use multiple learned queries for detecting various objects, our approach derives a single query directly from the natural language referring expression.

\vspace{-0.3cm}

\subsubsection{Cross-Attention Decoder Layers: }

Two stacked Transformer layers refine the text query through repeated cross-attention with the image features, progressively enriching the query representation with spatially-grounded visual information. Each layer performs cross-attention from the query to image patches, followed by a feed-forward layer and also with residual connections and layer normalization operations. Notice that the proposed architecture focuses computational resources on cross-modal attention, which directly contributes to spatial localization. Critically, each cross-attention operation produces attention weights representing a probability distribution over the $24 \times 24$ grid. These attention patterns can be used as the foundation for our bounding box prediction approach, as described in connection to the next stage.

\vspace{-0.3cm}

\subsubsection{Attention-Anchored Grounding: }

Our bounding box regression decomposes prediction into coarse anchor generation from spatial attention patterns, followed by learned fine-grained refinement. The considered anchor generation operation is non-differentiable, and thus we supervise attention patterns via a KL divergence, teaching the model to produce distributions that naturally yield accurate bounding box anchors. Cross-attention over the 24×24 grid produces normalized attention weights representing spatial confidence for each patch location. We consider the attention weights as a 2D distribution $A_{2D} \in \mathbb{R}^{24\times24}$ and marginalize to 1D distributions along the horizontal and vertical axes:
\begin{equation}
p_x(x) = \sum_{y=1}^{24} \mathbf{A}_{2D}(x,y), \quad p_y(y) = \sum_{x=1}^{24} \mathbf{A}_{2D}(x,y).
\end{equation}
\noindent After sharpening and re-normalizing the marginalized distributions through temperature scaling, we identify the spatial bounds of the predicted object by taking the 10th and 90th percentiles of the attention along each axis, capturing the central 80\% of the attention mass. The anchor box is then defined as $(x_1, y_1, x_2 - x_1, y_2 - y_1)$, where $x_1$ and $x_2$ correspond to the normalized coordinates for the 10th and 90th percentiles of the cumulative horizontal distribution $p_x$, and $y_1$ and $y_2$ similarly correspond to the 10th and 90th percentiles of the cumulative vertical distribution $p_y$. This percentile-based definition yields a tight, adaptive anchor that reflects the spatial spread of attention while remaining robust to outliers and diffuse activations.
The percentile level was fixed a priori so as to capture the central 80\% of the attention mass, and it was not tuned on validation data; a systematic sensitivity analysis over this value, and over the sharpening temperature, is left for future work. Notice that other procedures could also have been considered for generating the anchors (e.g., a bilinear interpolation of the attention weights into pixel coordinates, followed by greedy search to find the smallest bounding box that contains 80\% of the attention). The procedure based on percentiles is nonetheless simpler, and it can be extended to use differentiable relaxations of sorting when computing the percentiles \cite{grover2018stochastic, blondel2020fast}, which is an option that we plan to consider as future work. Still, the current anchor generation process is non-differentiable, and thus we guide it through attention supervision separately from fine-grained refinement, which in turn is learned via coordinate losses. Specifically, we use a two-layer MLP with rectifier activations to refine the anchors,  by predicting bounded residuals $\delta \in [-1,1]^4$. We apply residuals in logit space rather than coordinate space to ensure equal gradient magnitude regardless of anchor position:
\begin{equation}
{b}_{pred} = \sigma(\text{logit}(b_{anchor}) + {\delta}).
\end{equation}
In the previous expression, $\sigma$ denotes the logistic sigmoid function, and the function $\text{logit}(x) = \log(x/(1-x))$. This logit-space formulation unwraps $[0,1]$ coordinates to $(-\infty, \infty)$, providing stable gradients while constraining the final predictions to valid normalized coordinates. Unlike direct regression approaches that must learn full coordinate prediction from scratch, our method requires only learning residual adjustments over the attention-guided initial estimates.

\vspace{-0.3cm}

\subsubsection{Multi-Objective Loss Function: }

We train the model with three complementary loss components, weighted to balance coordinate accuracy, geometric overlap, and attention supervision:
\begin{equation}
\mathcal{L}_{total} = \lambda_{L1} \mathcal{L}_{L1} + \lambda_{GIoU} \mathcal{L}_{GIoU} + \lambda_{attn} \mathcal{L}_{attn}.
\end{equation}
The three different components have weights of $\lambda_{L1}=5.0$, $\lambda_{GIoU}=2.0$, and $\lambda_{attn}=0.2$. The first component is the standard smooth L1 loss between predicted and ground-truth coordinates, averaged over the batch, which provides stable coordinate regression without the outlier sensitivity of the more typical L2 loss, and which serves as the primary learning signal. The second component uses a generalized IoU (GIoU) score that directly optimizes geometric overlap between predicted and ground truth bounding boxes, by extending the standard IoU to provide meaningful gradients even for non-overlapping cases \cite{Rezatofighi_2018_CVPR}. It is defined as follows:
\begin{equation}
\text{GIoU} = \text{IoU} - \frac{|C \setminus (A \cup B)|}{|C|},
\end{equation}
where $A$ and $B$ denote the predicted and ground truth bounding boxes, and where $C$ is the smallest enclosing box covering both $A$ and $B$. The second term penalizes the normalized area outside the union of $A$ and $B$, encouraging the predicted bounding box to move closer to the ground truth even when the IoU is equal to 0. The corresponding loss is $\mathcal{L}_{GIoU} = 1 - \text{GIoU}$, providing a more stable optimization signal than the standard IoU. Finally, the attention supervision loss explicitly teaches the model where to look given a referring expression, supporting our attention-anchored prediction approach. We convert the ground-truth bounding box into a binary target map over the $24 \times 24$ grid by assigning $r_i = 1$ for patches inside the box and $r_i = 0$ otherwise. This binary mask is normalized to form the optimization target with a probability distribution $p_{\text{target}}$ over the 576 patches. The loss minimizes the KL divergence between predicted attention $a_{norm}$ and this target distribution:
\begin{equation}
\mathcal{L}_{attn} = \sum_{i=1}^{576} p_{target}(i) \log \frac{p_{target}(i)}{a_{norm}(i)}.
\end{equation}
The previous expression guides the model on where to look given a referring expression. Accurate attention patterns produce better anchor boxes, which in turn simplify the bounding box refinement task.

\begin{table}[!t]
\centering
\caption{Comparison of the attention-anchored model with state-of-the-art referring expression comprehension methods across multiple English benchmark datasets. The second column lists previously published models evaluated on each dataset, which we use as baselines for comparison. Results are reported in two configurations: Aggregate multilingual (A), combining all 10 languages, and English-only (E) evaluation. Best results for each metric are shown in bold, while dashes (–) indicate metrics not reported in the original publications. Baseline parameter counts are given where publicly reported, to contextualize the comparison against our 469M-parameter model (n/r: not reported by the authors; n/d: not disclosed, or not applicable because the system is a pipeline built on top of proprietary models).}
\label{tab:baseline_comparison}
\resizebox{\columnwidth}{!}{%
\begin{tabular}{l|l|c|cc|cc}
\toprule
\multirow{2}{*}{\textbf{Dataset}} & 
\multicolumn{1}{c|}{\multirow{2}{*}{\makecell{\textbf{Baseline English}\\\textbf{Method}}}} & 
\multicolumn{1}{c|}{\multirow{2}{*}{\makecell{\textbf{Baseline}\\\textbf{Params}}}} & 
\multicolumn{2}{c|}{\textbf{IoU@50}} & 
\multicolumn{2}{c}{\textbf{mIoU}} \\
& & & Baseline & Ours (A / E) & Baseline & Ours (A / E) \\
\midrule
FG-OVD & Detic \cite{bianchi2024devil, zhou2022detecting} & n/r & \textbf{0.697} & 0.512 / 0.565 & -- & 0.463 / 0.509 \\
\midrule
FineCops-Ref & CogVLM \cite{liu2024finecopsrefnewdatasettask} & 17B & \textbf{0.782} & 0.568 / 0.603 & -- & 0.520 / \textbf{0.550} \\
\midrule
LocateBench & GPT-4o \cite{chiang2024locatebenchevaluatinglocatingability} & n/d & 0.604 & 0.552 / \textbf{0.614} & -- & 0.528 / \textbf{0.577} \\
\midrule
Multimodal Ground & MRVG-NET \cite{lu2025multimodalreferencevisualgrounding} & n/d & \textbf{0.807} & 0.544 / 0.532 & \textbf{0.798} & 0.458 / 0.489 \\
\midrule
RefCLEF & LMSVA \cite{LMSVA-YAO2024105242} & n/r & \textbf{0.730} & 0.618 / 0.681 & -- & 0.565 / \textbf{0.612} \\
\midrule
\multirow{6}{*}{RefCOCO} 
  & CogVLM \cite{Cogvlm-wang2024cogvlm}    & 17B & \textbf{0.920} & \multirow{6}{*}{0.869 / 0.913} & \multirow{6}{*}{--} & \multirow{6}{*}{0.788 / \textbf{0.822}} \\
  & Grounding DINO \cite{liu2023groundingdinomarryingdino} & 341M & 0.908 & & & \\
  & OFA-Large \cite{wang2022ofa}            & 472M & 0.893 & & & \\
  & Qwen-VL \cite{Qwen-VL}              & 9.6B & 0.886 & & & \\
  & UNINEXT-H \cite{uninext}           & 673M & 0.824 & & & \\
\midrule
\multirow{6}{*}{RefCOCO+} 
  & CogVLM    & 17B & \textbf{0.885} & \multirow{6}{*}{0.689 / 0.746} & \multirow{6}{*}{--} & \multirow{6}{*}{0.637 / 0.634} \\
  & Grounding DINO     & 341M & 0.827 & & & \\
  & OFA-Large            & 472M & 0.849 & & & \\
  & Qwen-VL               & 9.6B & 0.831 & & & \\
  & UNINEXT-H            & 673M & 0.712 & & & \\
\midrule
RefCOCOg & CogVLM & 17B & \textbf{0.897} & 0.765 / 0.793 & -- & 0.699 / 0.720 \\
\midrule
RefDrone & NGDINO-T \cite{sun2025refdronechallengingbenchmarkreferring} & $\sim$172M & 0.218 & 0.236 / \textbf{0.257} & -- & 0.225 / \textbf{0.241} \\
\midrule
Toloka & wztxy89 \cite{toloka-gao2023champion} & n/r & \textbf{0.834} & 0.583 / 0.626 & \textbf{0.763} & 0.502 / 0.535 \\
\midrule
\rowcolor{gray!15}
\textbf{Overall} & -- & -- & -- & \textbf{0.625 / 0.672} & -- & \textbf{0.571 / 0.609} \\
\bottomrule
\end{tabular}%
}
\end{table}
\vspace{-0.25cm}

\section{Experimental Evaluation}
\label{sec:resul}

\vspace{-0.25cm}

We evaluated the proposed attention-anchored model across three different main dimensions: aggregate performance on the complete multilingual test set, dataset-specific analysis to assess coverage of different domains, and per-language assessment quantifying multilingual consistency. Additional experimental results are presented in the supplementary materials.

Training was made on a single NVIDIA RTX 4090 with batch size 32 and mixed precision, running for 4 epochs over 6.3M multilingual expression-image pairs. Each epoch required approximately 26 GPU-hours, for a total training time of approximately 104 hours. The model has 469M total parameters, with the majority of the SigLIP2 backbone remaining frozen -- only the final three layers of each encoder, the cross-modal fusion, and the task-specific components are trained. In total, approximately 35M of the 469M parameters are trainable. We used AdamW with weight decay of 0.01 and gradient clipping with norm 1.0 \cite{loshchilov2018decoupled}. Learning rates are $10^{-6}$ for the SigLIP2 encoder’s top layers, $10^{-5}$ for cross-modal fusion, and $10^{-4}$ for the task-specific components. Evaluation covers 429,154 referring expressions across all test splits. Metrics include mean IoU (mIoU) and IoU@50, being computed globally, per dataset, and per language, to examine localization quality and multilingual generalization.

\begin{figure}[!t]
    \centering
    \includegraphics[width=0.75\linewidth]{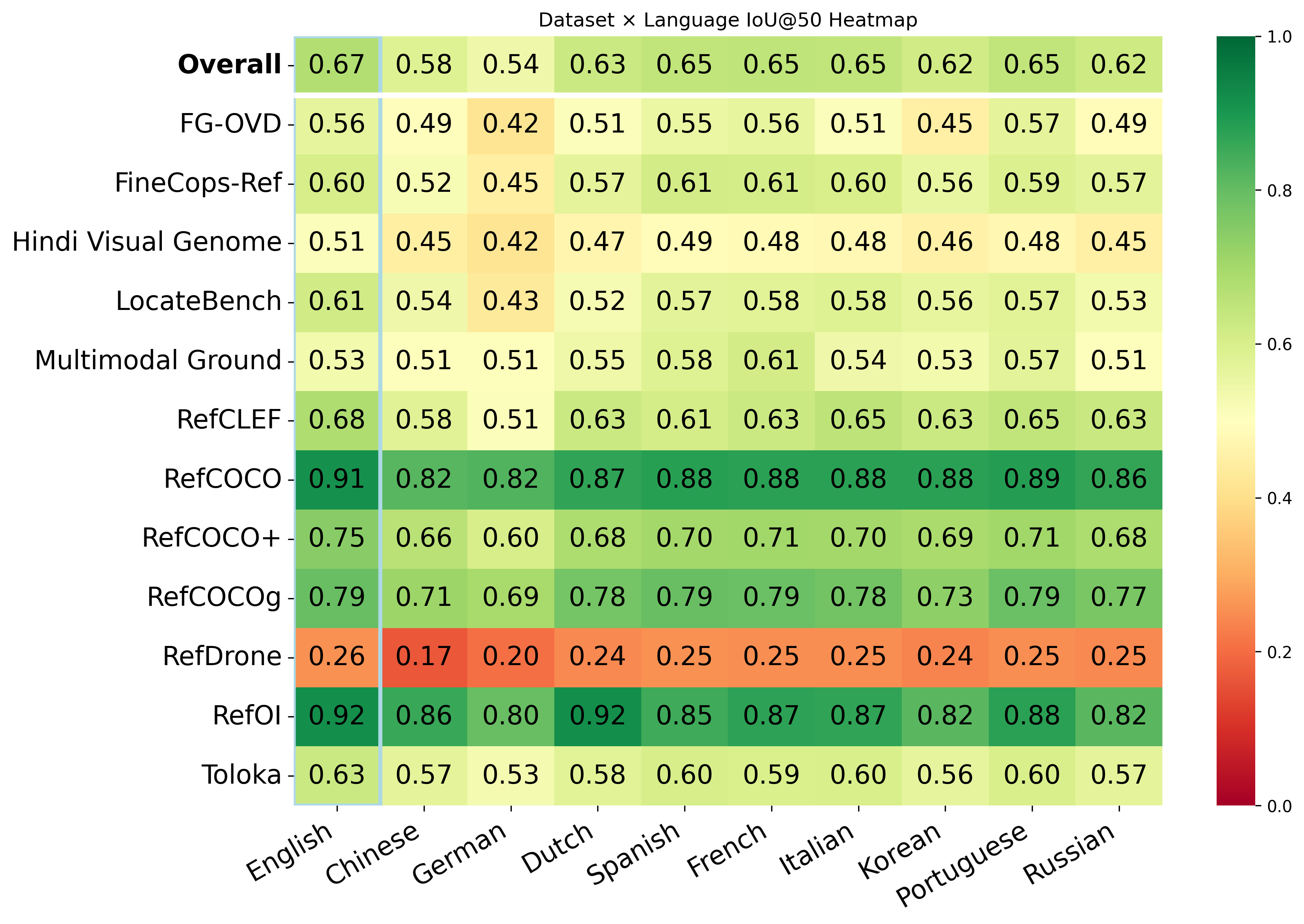}
    \caption{Multilingual performance in terms of IoU@50 across datasets and languages. Rows represent datasets, while columns represent languages.}
    \label{fig:PerfLangHeatmap}
\end{figure}


\vspace{-0.25cm}

\subsection{Aggregate Performance}

\vspace{-0.25cm}

We evaluated the attention-anchored model on the 429,154 test expressions from the complete dataset. Several of the sub-datasets nonetheless require clarifications. FG-OVD and RefCOCOg lack official test splits, so we reserved half of their validation sets for testing. Hindi Visual Genome and RefOI were excluded from the detailed presentation in Table~\ref{tab:baseline_comparison} due to absence of published REC baselines and methodological inconsistencies with multi-object evaluation protocols, respectively. For RefCOCO and RefCOCO+, we did not distinguish between the original testA and testB splits, and instead merged them into a single test set for evaluation. To enable a fair comparison with previous studies that report separate results for each split, we adjusted the reported accuracies using a weighted average based on the number of samples in each test subset. Finally, the RefCOCOg split was compared only for its reported validation score.

Table~\ref{tab:baseline_comparison} presents the overall results, together with a comparative performance assessment for some of the sub-datasets against previously published state-of-the-art results (often corresponding to much larger LLMs, or models finetuned specifically to one single dataset), reporting evaluation metrics in two configurations, namely Aggregate multilingual (A) and English-only (E). In the aggregate multilingual setting, the proposed model achieves a mean IoU of 0.571 and a median of 0.721 across the complete test set. Accuracy metrics show that 62.5\% of the predictions exceed the IoU threshold of 0.5, while 50.1\% exceed the stricter threshold of 0.7. The gap between the mean (0.571) and the median (0.721) indicates a right-skewed score distribution.

\vspace{-0.25cm}

\subsection{Dataset-Level Performance}
Performance varies substantially across the original English datasets, as detailed per dataset in Tables~\ref{tab:baseline_comparison} and~\ref{tab:ablation_comparison}. The RefCOCO family gives the strongest results, with RefCOCO reaching 0.788 mean IoU (86.9\% IoU@50, 80.8\% IoU@70) and RefCOCO+ reaching 0.637 mean IoU (68.9\% IoU@50, 61.9\% IoU@70). At the other extreme, RefDrone is markedly harder, at 0.225 mean IoU (23.6\% IoU@50) with a median IoU of only 0.022, reflecting differences associated to the aerial viewpoint and the prevalence of small targets.

Compared to published baselines, our model achieves competitive performance. For instance, on RefCOCO, we surpass the Grounding Dino \cite{liu2023groundingdinomarryingdino} baseline and perform similarly to the much larger CogVLM model \cite{Cogvlm-wang2024cogvlm}. For RefDrone, we improve over the NGDINO-T baseline \cite{sun2025refdronechallengingbenchmarkreferring}, demonstrating progress on the challenging aerial domain despite the absolute performance remaining low. However, significant gaps persist on Toloka VQA (83.4\% for the best system in the competition vs.\ 62.6\% for ours in the English-only setting) and Multimodal Ground (80.7\% vs.\ 53.2\%). On LocateBench, our English-only result slightly exceeds the GPT-4o baseline (61.4\% vs.\ 60.4\%), validating effectiveness on multiple-choice localization, despite the different task format. It is important to note that our evaluation on RefOI employed the single-occurrence mode exclusively. Consequently, direct quantitative comparison with published results is not feasible, as the original benchmarks incorporate both single-occurrence and multi-occurrence evaluation protocols.

\begin{table}[!t]
\centering
\setlength{\tabcolsep}{3pt}
\renewcommand{\arraystretch}{1.0}
\tiny
\begin{minipage}[t]{0.56\linewidth}
\centering
\captionof{table}{Ablation study comparing the attention-anchored approach against direct regression (IoU@50). Results are reported for aggregate multilingual (A) and English-only (E) evaluation. The best results are in \textbf{bold}.}
\label{tab:ablation_comparison}
\end{minipage}%
\hfill
\begin{minipage}[t]{0.42\linewidth}
\centering
\captionof{table}{English-only vs.\ multilingual training (IoU@50). Both models use the same architecture and combined dataset; only training language differs. Best results in \textbf{bold}.}
\label{tab:multilingual_value}
\end{minipage}

\vspace{4pt}

\begin{minipage}[c]{0.56\linewidth}
\centering
\begin{tabular}{l|cc|cc|cc}
\toprule
& \multicolumn{2}{c|}{Attn-Anch.} & \multicolumn{2}{c|}{Direct-Reg.} & \multicolumn{2}{c}{$\Delta$} \\
Dataset & A & E & A & E & A & E \\
\midrule
RefCOCO            & \textbf{86.9} & \textbf{91.3} & 79.4 & 87.7 & +7.5  & +3.6  \\
RefCOCO+           & \textbf{68.9} & \textbf{74.6} & 57.8 & 62.9 & +11.1 & +11.7 \\
RefCOCOg           & \textbf{76.5} & \textbf{79.3} & 65.2 & 69.3 & +11.3 & +10.0 \\
RefCLEF            & \textbf{61.8} & \textbf{68.1} & 56.8 & 62.8 & +5.0  & +5.3  \\
RefDrone           & \textbf{23.6} & \textbf{25.7} & 7.5  & 8.8  & +16.1 & +16.9 \\
RefOI              & 85.8          & \textbf{91.8} & \textbf{86.3} & 88.0 & -0.5  & +3.8  \\
Toloka             & \textbf{58.3} & \textbf{62.6} & 45.4 & 48.6 & +12.9 & +14.0 \\
LocateBench        & \textbf{55.2} & \textbf{61.4} & 41.4 & 44.6 & +13.8 & +16.8 \\
FineCops-Ref       & \textbf{56.8} & \textbf{60.3} & 48.3 & 51.8 & +8.5  & +8.5  \\
Multimodal Ground  & \textbf{54.4} & \textbf{53.2} & 21.6 & 23.4 & +32.8 & +29.8 \\
Hindi VG           & \textbf{46.7} & \textbf{50.8} & 41.8 & 46.8 & +4.9  & +4.0  \\
FG-OVD             & \textbf{51.2} & \textbf{56.5} & 36.9 & 42.4 & +14.3 & +14.1 \\
\midrule
\rowcolor{gray!15}
Overall            & 62.5 & 67.2 & 54.2 & 59.2 & +11.3 & +8.0 \\ 
\bottomrule
\end{tabular}
\end{minipage}%
\hfill
\begin{minipage}[c]{0.42\linewidth}
\centering
\begin{tabular}{l|cc|c}
\toprule
Language & EN-Only & Multi. & $\Delta$ \\
\midrule
English    & 52.1 & \textbf{67.2} & +15.1 \\
Spanish    & 36.6 & \textbf{64.6} & +28.0 \\
French     & 31.0 & \textbf{64.8} & +33.8 \\
Italian    & 30.2 & \textbf{64.5} & +34.3 \\
Portuguese & 33.6 & \textbf{64.5} & +30.9 \\
Dutch      & 29.6 & \textbf{62.7} & +33.1 \\
Russian    & 27.1 & \textbf{62.2} & +35.1 \\
Korean     & 23.7 & \textbf{61.5} & +37.8 \\
Chinese    & 30.2 & \textbf{58.3} & +28.1 \\
German     & 21.2 & \textbf{54.1} & +32.9 \\
\midrule
\rowcolor{gray!15}
\textbf{Overall} & 31.5 & \textbf{62.6} & \textbf{+31.1} \\
\bottomrule
\end{tabular}
\end{minipage}
\end{table}

\vspace{-0.25cm}

\subsection{Multilingual Performance}

\vspace{-0.25cm}

Figure~\ref{fig:PerfLangHeatmap} presents per-language performance across all the datasets using IoU@50 as the evaluation metric. English establishes the performance ceiling at 63.8\% IoU@50, with other languages exhibiting relatively modest degradation. Romance languages cluster within a narrow band, with Spanish (61.3\% IoU@50), French (61.4\% IoU@50), Italian (61.3\% IoU@50), and Portuguese (61.3\% IoU@50) achieving near-identical performance, differing by at most 0.1 percentage points from the English results. Germanic languages show greater variation, with Dutch achieving 59.2\% IoU@50, and German reaching 50.8\% IoU@50. Russian obtains 59.1\% IoU@50, while East Asian languages show 58.7\% IoU@50 for Korean, and 55.5\% IoU@50 for Chinese. Performance gaps between English and other languages thus range from 2.5 percentage points (Italian: 63.8\% vs. 61.3\%) to 13 percentage points (German: 63.8\% vs. 50.8\%) in terms of the IoU@50 metric.

\vspace{-0.25cm}

\subsection{Ablation Experiments}

\vspace{-0.25cm}

To validate the contribution of the attention-anchored prediction mechanism, we compared our complete approach against a baseline variant that uses direct regression, to isolate the impact of anchor generation and residual refinement. Both models share the same cross-attention decoder, differing only in the final prediction head. In a separate ablation study, we compared the proposed model when trained on the combined dataset using only English expressions, versus the full multilingual training. Notice that even when not using multilingual training data our model can still process multilingual expressions, due to the use of natively multilingual SigLIP2 encoder.

Table~\ref{tab:ablation_comparison} reports the IoU@50 score across the 12 sub-datasets and also overall, showing that the attention-anchored approach consistently outperforms direct regression, with overall gains of +11.3 percentage points in the multilingual evaluation, and of +8.0 in the English-only evaluation (the mean gain across individual datasets is larger, at approximately +11.5 percentage points in both settings). Improvements are more pronounced on datasets with small or visually complex targets, such as Multimodal Ground (+32.8). RefOI is the single case where direct regression slightly surpasses anchored prediction in terms of the aggregate evaluation (-0.5), although the anchored approach remains superior in the English-only evaluation. 

In turn, Table~\ref{tab:multilingual_value} shows that multilingual training nearly doubles overall performance (31.5\%$\rightarrow$62.6\%, +98.7\% relative improvement), and that English performance also improves under multilingual training (+15.1 percentage points).

\vspace{-0.25cm}

\section{Conclusions and Future Work}
\label{sec:discussion}

\vspace{-0.25cm}

This work established foundational infrastructure for multilingual Referring Expression Comprehension (REC) through two primary contributions: (1) a unified multilingual dataset expanding 12 English REC benchmarks across 10 languages through systematic translation, quality validation, and visual context enhancement, yielding 8M multilingual referring expressions; and (2) an attention-anchored neural architecture that decomposes localization into coarse spatial attention and learned refinement, achieving language-agnostic grounding through frozen multilingual encoders and interpretable attention mechanisms. 

Despite its small size, we have that the performance of the proposed model is competitive over multiple individual English-centric baselines, while achieving generalization to multiple languages and domains, validating the feasibility of efficient multilingual REC methods that can serve users in their native languages.

Important limitations include (1) the non-differentiable anchor generation procedure requiring explicit attention supervision via KL divergence, potentially limiting any extensions towards multi-object scenarios, (2) spatial resolution constraints from the 16$\times$16 pixel patch granularity, preventing sub-patch localization precision for small objects, and (3) the single-object prediction constraint, not abstaining from generating bounding boxes for expressions that are not grounded in the image~\cite{he2023grec,yang2026teaching}, or having that plural expressions referencing multiple instances result in selecting only one target encompassing bounding box. Future work can explore multi-scale feature extraction for improved small object handling, as well as alternative anchor generation methods as discussed in Section~\ref{sec:methodology}. We also plan to extend the dataset to additional languages and specialized domains (e.g., medical or satellite imagery).


\section*{Acknowledgements}

This research was supported by the Portuguese Recovery and Resilience Plan through project C645008882-00000055 (i.e., the Center For Responsible AI), and also by the Funda\c{c}\~ao para a Ci\^encia e a Tecnologia, I.P. (FCT) under the projects with references 2024.14126.UTA (project FOMO-HODOR), 2024.12756.CMU (project Bomox), UID/50021/2025, UID/PRR/50021/2025, LA/P/0083/2020, UIDP/50009/2020 and UIDB/50009/2020.

\section*{Data, Code, and Model Availability}

The unified multilingual dataset, the trained model weights, and the source code used for dataset construction, training, and evaluation are publicly available at \url{https://multilingual.franreno.com}. We specifically release the translated referring expressions together with the bounding box annotations, quality scores, and dataset provenance metadata. The underlying images remain subject to the licenses of their original source datasets, to which we provide pointers rather than redistributing image content.

%
%
\bibliographystyle{splncs04}
\bibliography{main}

\section*{Appendix A: Limitations and Ethical Considerations}
\label{sec:appA}

While our model aims to contribute to research beyond English-centric referring expression comprehension, it has limitations in that the results are only conditioned on, and evaluated on, machine translated data from existing English-centric datasets. This fundamental constraint means that despite our efforts to enhance translation quality through visual context-based refinement, the multilingual dataset inherits potential biases and limitations from both the original English-language sources and the translation process itself. Machine translation, even when enhanced with multimodal grounding, cannot fully capture the nuanced ways different linguistic communities naturally describe visual scenes, as referring expression conventions reflect cultural and linguistic patterns that may not transfer directly through translation.

The data used in our experiments is also limited in the coverage of geographically diverse concepts, reflecting the predominantly Western-centric nature of the underlying image collections (i.e., MS COCO, Visual Genome, or OpenImages). Objects, scenes, and spatial arrangements common in other global contexts may be underrepresented or absent entirely, potentially limiting the model's effectiveness when deployed in diverse real-world settings. Even our visual context-based refinement procedure, while improving translation quality measurably, is not fully capable of avoiding problems inherent to automatic machine translation, such as loss of pragmatic meaning, cultural connotations, or language-specific spatial reasoning patterns documented in linguistic literature.

Future work in this area should invest further in the annotation and curation of truly multilingual data for referring expression comprehension, ideally collected from native speakers describing images relevant to their own cultural contexts. This would involve expanding the range of covered languages beyond our current ten-language scope, particularly to include low-resource languages and linguistic families not represented in our dataset (e.g., African, Indigenous, and Southeast Asian languages). Additionally, research should explore how referring expression conventions vary across culture, and whether vision-language models can learn these language-specific patterns rather than imposing English-centric spatial reasoning universally.

\section*{Appendix B: Additional Discussion on the Results}
\label{sec:appB}

We further discuss here the per-dataset performance and the multilingual results.

\subsection*{Appendix B1: Per-Dataset Analysis}
\label{sec:appB1}

The obtained results demonstrate a competitive performance on standard REC benchmarks. For instance, RefCOCO achieves an average result of 86.9\% IoU@50, and RefCOCO+ obtains 68.9\% IoU@50. The strong performance on the RefCOCO family validates that our attention-anchored approach effectively handles diverse referring expression styles and complex scenes with multiple objects. Moreover, the gap between mean (0.571) and median (0.721) IoU reflects a right-skewed distribution where most predictions achieve strong localization, while domain-specific challenges contribute to lowering the aggregate mean.

It is interesting to note that previous studies on Referring Expression Comprehension (REC) typically evaluate performance separately on two distinct test splits for RefCOCO and RefCOCO+, commonly referred to as testA and testB. In turn, for RefCOCOg, standard evaluation protocols often employ the complete validation set. To facilitate a more direct comparison with prior work, and to provide comprehensive evaluations aligned with established benchmarks, we report our model's performance following these evaluation protocols in Table~\ref{tab:refcoco_splits}.

\begin{table*}[!t]
\caption{English-only evaluation on standard RefCOCO family benchmark splits. Previous work typically reports performance separately for the testA and testB splits of RefCOCO and RefCOCO+, and uses the full validation set for RefCOCOg evaluation. Results are reported using the IoU@50 metric. Best baseline results are shown in \textbf{bold}.}
\label{tab:refcoco_splits}
\centering
\begin{tabular}{l|c|cc|cc|c}
\toprule
\multirow{2}{*}{\textbf{Model}} 
& \multirow{2}{*}{\textbf{Parameters}}
& \multicolumn{2}{c|}{\textbf{~RefCOCO~}} 
& \multicolumn{2}{c|}{\textbf{~RefCOCO+~}}
& \textbf{RefCOCOg} \\
& & \textbf{~test-A~} & \textbf{~test-B~}
& \textbf{~test-A~} & \textbf{~test-B~}
& \textbf{val} \\
\hline
CogVLM & 17B & \textbf{94.75} & \textbf{88.99} & \textbf{92.91} & \textbf{83.39} & \textbf{89.75} \\
\hline
Grounding DINO & 341M & 93.19 & 88.24 & 88.95 & 75.92 & 87.02 \\
\hline
OFA-Large & 472M & 83.67 & 76.39 & 76.00 & 61.75 & 67.57 \\
\hline
Qwen-VL & 9.6B & 92.26 & 85.34 & 88.25 & 77.21 & 85.58 \\
\hline
UNINEXT-H & 673M & 94.33 & 91.46 & 89.63 & 79.79 & 88.73 \\
\hline
\rowcolor{gray!15}
Ours & 469M & 90.61 & 84.42 & 81.93 & 66.22 & 76.42 \\
\bottomrule
\end{tabular}
\end{table*}

Our implementation demonstrates competitive performance across all splits, though we observe performance variations aligned with the dataset split characteristics. Results with our generalist model are indeed slightly inferior, but notice that the comparison is being made against much larger generalist vision-and-language models (e.g., CogVLM or Qwen-VL), or against specialized models individually finetuned in each dataset (e.g., Grounding DINO). On RefCOCO, we achieve 90.61\% IoU@50 on testA and 84.42\% on testB. RefCOCO+ exhibits a more pronounced performance differential, particularly on testB (66.22\% vs. 83.39\% for CogVLM, a 17.2 point gap). RefCOCOg validation achieves 76.42\% IoU@50, maintaining consistency with our previously reported aggregate results.

RefDrone's aerial imagery presents the most significant challenge (0.225 mean IoU), likely due to three factors: (1) non-standard overhead viewpoints that differ from the predominantly ground-level perspectives in the SigLIP2 pre-training data; (2) small object scales (31.1\% small targets) that occupy few patches in our 24×24 grid, making percentile-based anchor generation less reliable; (3) occlusions and viewing angle variations that create ambiguous spatial relationships. These results suggest future work should incorporate multi-scale feature extraction or adaptive grid resolutions for small object handling.

\subsection*{Appendix B2: Comparison Against Zero-Shot Use of Qwen3-VL-8B}
\label{sec:appB2}

To further contextualise the contribution of our specialised multilingual approach, we also compared it against the use of the Qwen3-VL-8B-Instruct model in a zero-shot setting, on the test split of our multilingual REC benchmark. The model receives the image and the referring expression in its native language, with no task-specific training, and is prompted to return a normalised bounding box in \texttt{[x\_min, y\_min, x\_max, y\_max]} format. The prompt that was used is as follows:

\medskip
\noindent\parbox{\linewidth}{\small\ttfamily\raggedright\sloppy
Given the image, locate the object described by the following expression and return the bounding box coordinates. Expression: \{expression\} Return the bounding box in the format: [x\_min, y\_min, x\_max, y\_max] where coordinates are normalized to [0, 1] range (0 is top-left, 1 is bottom-right). IMPORTANT: Only return the four numbers in brackets, nothing else. Example: [0.25, 0.30, 0.75, 0.80]
}

Notice that the Qwen3-VL-8B-Instruct model is natively multilingual, although it was not specifically trained for performing referring expression comprehension from multilingual expressions. This model is also much larger than the attention-guided model advanced in our work, and although even larger models could have been considered for the comparison, Qwen3-VL-8B-Instruct represents state-of-the-art results from a generalist model that can still be executed on relatively modest hardware.

\begin{table}[t]
\centering
\caption{Zero-shot referring expression comprehension results for Qwen3-VL-8B-Instruct on our multilingual benchmark test split, evaluated with IoU@50.}
\label{tab:qwen3vl_zeroshot}
\begin{tabular}{l|c|c}
\toprule
\textbf{Language~~} & \textbf{~Qwen3-VL-8B~} & \textbf{~Ours (469M)~} \\
\midrule
English     & 63.5 & \textbf{67.2} \\
Spanish     & 62.1 & \textbf{64.6} \\
French      & 62.7 & \textbf{64.8} \\
Italian     & 61.7 & \textbf{64.5} \\
Portuguese  & 61.8 & \textbf{64.5} \\
Dutch       & 59.9 & \textbf{62.7} \\
Russian     & 61.9 & \textbf{62.2} \\
Korean      & 59.5 & \textbf{61.5} \\
Chinese     & 59.6 & \textbf{58.3} \\
German      & \textbf{61.9} & 54.1 \\
\midrule
\textbf{Average} & \textbf{61.5} & \textbf{62.5} \\
\bottomrule
\end{tabular}
\end{table}

Table~\ref{tab:qwen3vl_zeroshot} reports per-language IoU@50 and mIoU results. The Qwen3-VL-8B model achieves an overall IoU@50 of 61.5\% and a mean IoU of 0.561 averaged across all ten languages, which is notably below the performance of our model, despite Qwen3-VL-8B having approximately 17$\times$ more parameters. This performance gap shows that strong general-purpose visual language understanding does not straightforwardly translate into accurate multilingual referring expression grounding, and underscores the value of task-specific multilingual training.

Among the different languages, English achieves the strongest result (63.5\% IoU@50), consistent with the English-centric nature of most MLLM pretraining corpora. Non-Latin script languages show the greatest degradation: Korean (59.5\%) and Chinese (59.6\%) rank among the weakest, a pattern attributable to both reduced multilingual supervision in the base model and the increased tokenisation complexity of ideographic and syllabic scripts. Dutch (59.9\%) and German (61.9\%) show a small but consistent underperformance relative to Romance languages, which cluster tightly between 61.7\% and 62.7\%, suggesting that the model's multilingual grounding capacity broadly tracks the language distribution of its pretraining data.

\section*{Appendix C: Additional Details About the Translation Enhancement Methodology}
\label{sec:appC}

This appendix provides complete implementation details for the visual context-based translation enhancement pipeline, which elevated baseline translation quality across all target languages. We present the full prompt template employed with Google's Gemini 2.0 Flash Lite model, and we also analyze the resulting quality improvements.

\subsection*{C.1: The Prompt Template that Uses the Visual Context}
\label{sec:appC1}

The translation enhancement pipeline employed the following prompt template with Google's Gemini 2.0 Flash Lite model. Each instance received three inputs: the original English expression, the initial text-only translation, and the source image with a target object bounding box overlay.

\begin{verbatim}
You are a translation-quality-improvement assistant.

Original Text in English: {expression_text}

Translation to {language_code}: {translation}

Instruction: Given the original text and its translation, and 
taking the image as context, improve the quality of the 
translation by rephrasing it. Ensure the rephrased translation 
closely aligns with the original text in meaning, structure, tone, 
and style. Make the rephrased translation sound natural and fluent 
in the target language while preserving all essential details, 
correcting any grammatical errors, and retaining all stylistic 
elements. The rephrased translation should not be ambiguous, 
avoiding terms that can have a double meaning in the target 
language, or qualifying these terms with sufficient details for 
disambiguation.
\end{verbatim}

\subsection*{C.2: Improvements on the Distribution of Quality Scores}
\label{sec:appC2}

\begin{figure*}[!t]
    \centering
    \includegraphics[width=1\linewidth]{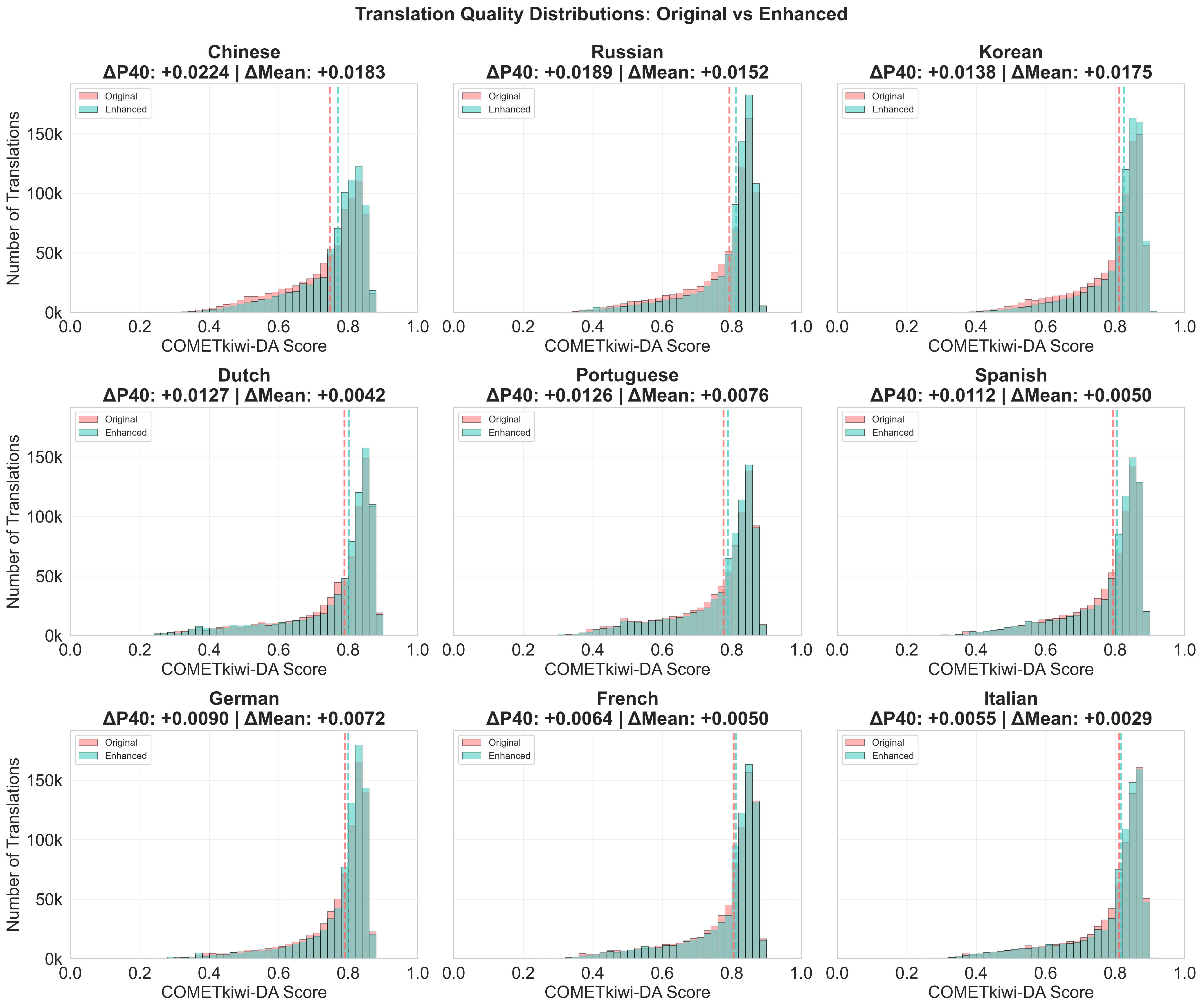}
    \caption{Translation quality distributions before and after visual enhancement for the nine target languages. Each panel shows overlaid histograms of COMETkiwi-DA scores for original (coral) and enhanced (teal) translations. All languages show rightward distribution shifts, with Chinese exhibiting the most substantial improvement.}
    \label{fig:cometkiwi_scores}
\end{figure*}

Figure~\ref{fig:cometkiwi_scores} presents the distribution of COMETkiwi-DA scores before and after visual context-based enhancement across the nine target languages. The histograms reveal consistent rightward shifts across all languages, indicating systematic quality improvements through visual enhancement. Chinese demonstrates the most substantial distributional shift, with the enhanced distribution (teal) showing marked concentration in higher quality ranges compared to the baseline (coral). This improvement is particularly significant given Chinese's initial lower performance ($P_{40}$ = 0.748), where 40.9\% of the baseline translations scored below 0.75. The $P_{40}$ thresholds, marked by vertical dashed lines, visually capture the magnitude of improvement, with Chinese showing the largest gap between baseline and enhanced thresholds, followed by Portuguese and Russian.

The language family patterns that were observed in terms of the baseline quality assessments persist through the enhancement, though with notable convergence. Romance languages (i.e., Spanish, French, Italian, and Portuguese) maintain their quality clustering while achieving consistent improvements, with all four languages shifting their distributions rightward. Germanic languages (i.e., German, and Dutch) and Korean, which exhibited stronger baseline performance, show more modest but meaningful shifts, suggesting that visual enhancement benefits lower-performing translations disproportionately. The reduced left-tail mass across all distributions indicates that enhancement particularly addresses the 26.5\% of baseline translations scoring below 0.75, elevating problematic translations toward acceptable quality thresholds, while maintaining the integrity of already high-quality translations.

\section*{Appendix D: Qualitative Analysis}
\label{sec:appD}

This appendix provides visual evidence for successful or problematic referring expression comprehension localizations, demonstrating the model's capabilities across diverse linguistic and visual scenarios. We present a collection of figures that illustrate both the strengths and limitations of the attention-anchored architecture through carefully selected examples, spanning multilingual consistency, compositional reasoning patterns, attribute-based discrimination, and particularly challenging failure modes.

\begin{figure*}[!t]
    \centering
    \includegraphics[width=1\linewidth]{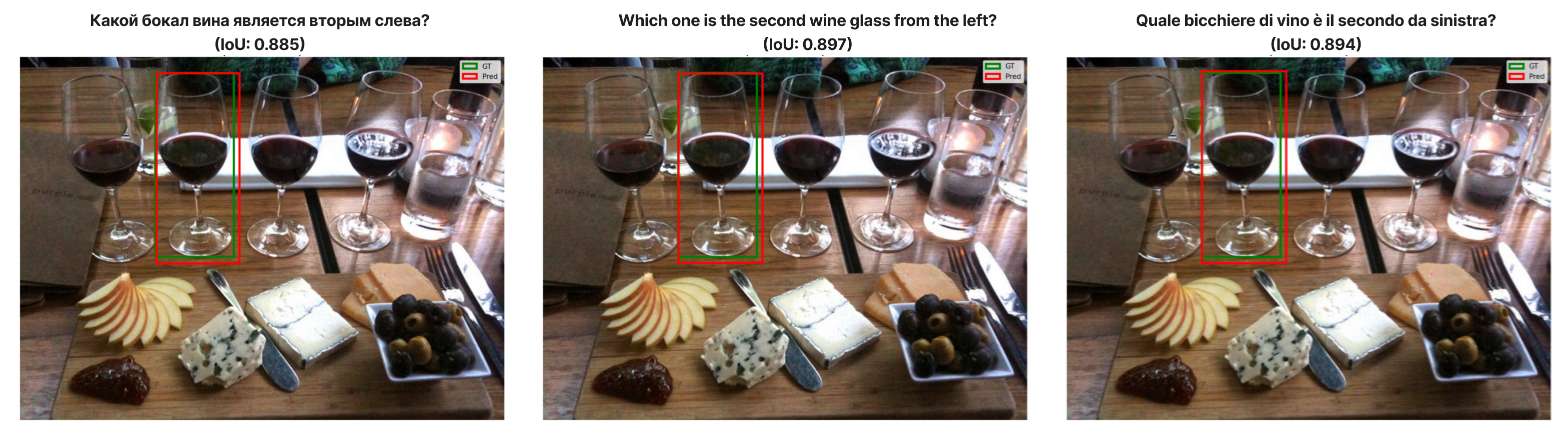}
    \caption{Multilingual localization consistency demonstrated on a wine tasting scene. The model accurately identifies the second wine glass from the left when queried in Russian (left), English (center), and Italian (right). Despite linguistic variations across three language families (Slavic, Germanic, and Romance), the spatial predictions remain nearly identical, with green boxes indicating the ground truth and red boxes showing the model predictions.}
    \label{fig:multilingual_wines}
\end{figure*}

\begin{figure*}
    \centering
    \includegraphics[width=0.9\linewidth]{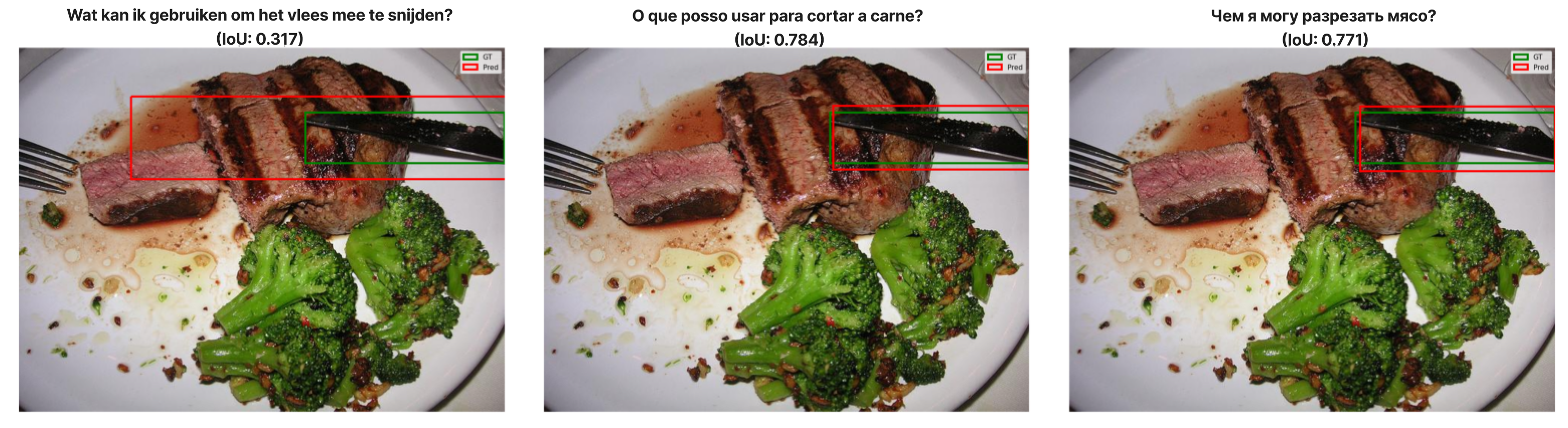}
    \caption{Cross-lingual localization consistency for a knife in a plated dish. The model produces spatially consistent predictions across Dutch (left), Portuguese (middle), and Russian (right) referring expressions, all targeting the same knife. Red boxes indicate the predictions, while green boxes show the ground truth. The slight deviation in the Dutch prediction (IoU: 0.784) compared to Portuguese (IoU: 0.771) and Russian (IoU: 0.802) may reflect subtle differences in translation quality or tokenization patterns, but all three achieve strong localization accuracy.}
    \label{fig:carnes}
\end{figure*}

\begin{figure*}
    \centering
    \includegraphics[width=0.9\linewidth]{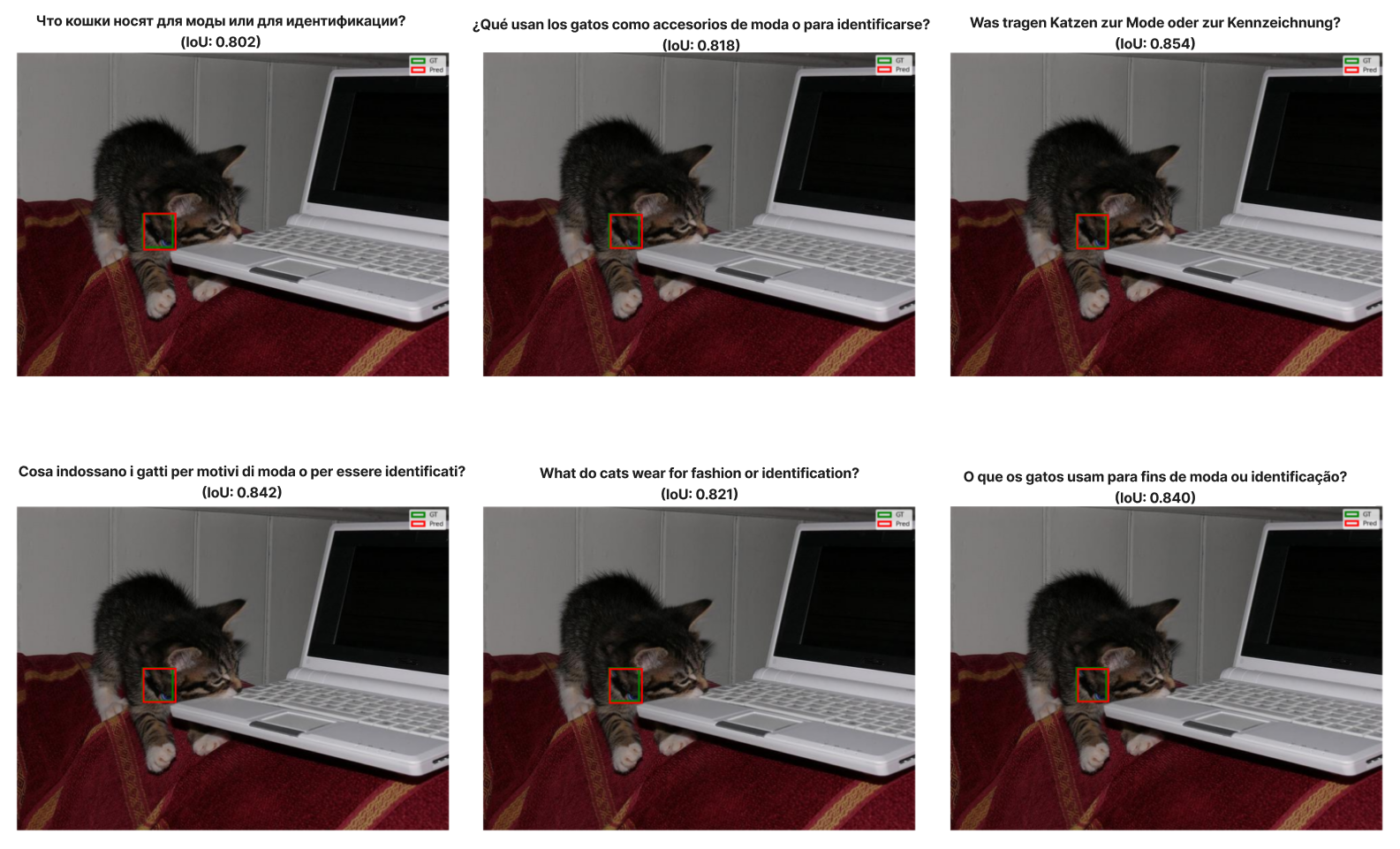}
    \caption{Illustration for small object localization across six languages: Russian, Portuguese, German, Italian, English, and Spanish. The model successfully identifies a cat's collar with consistent accuracy across all linguistic variants. The referring expressions all translate to "What do cats wear for fashion or identification?" from the respective languages. The mean IoU across languages is 0.831, demonstrating that small object localization accuracy remains stable in multilingual contexts.}
    \label{fig:smallobjectincat}
\end{figure*}

Multilingual consistency is demonstrated in Figures \ref{fig:multilingual_wines}, \ref{fig:carnes}, and \ref{fig:smallobjectincat}, showing that spatial predictions remain stable across language families, despite structural differences in referring expressions. For instance, Figure \ref{fig:multilingual_wines} presents a wine tasting scene where the model accurately localizes the second wine glass from the left when queried in Russian, English, and Italian, spanning three distinct language families (Slavic, Germanic, and Romance), with nearly identical spatial predictions across all three languages. This cross-lingual stability extends to various object categories and scene complexities, validating the language-agnostic nature of the learned spatial representations.

\begin{figure*}[!t]
    \centering
    \includegraphics[width=1\linewidth]{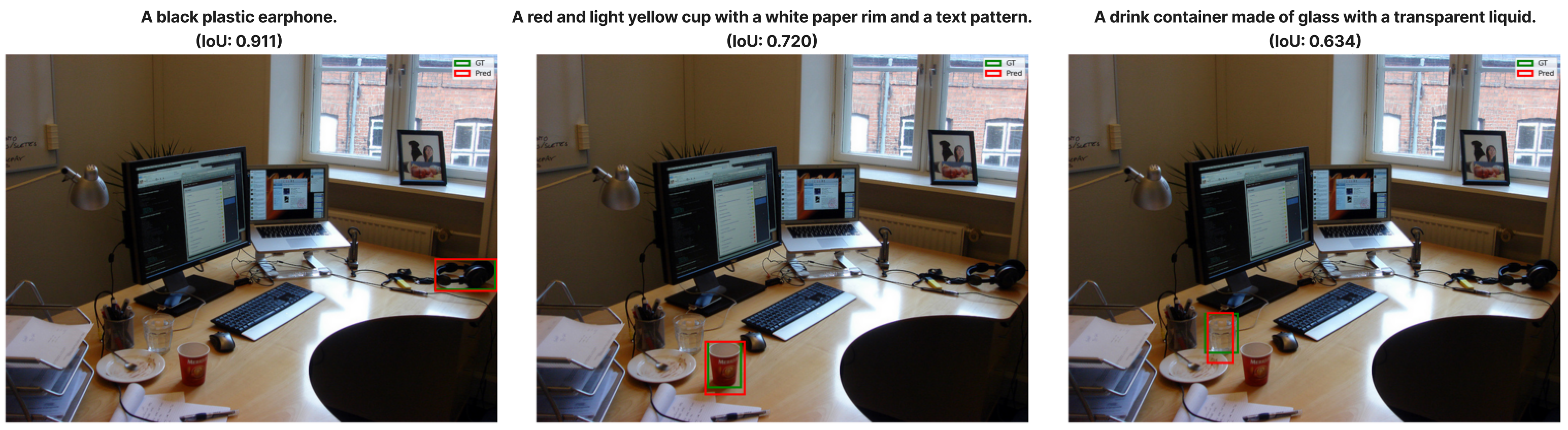}
    \caption{Examples illustrating fine-grained object discrimination in a multi-object scene. Three referring expressions targeting different objects in the same office environment demonstrate compositional reasoning: identifying an earphone by material and color (left), a cup by color and text pattern (center), and a glass container by material properties (right). The model successfully disambiguates targets from distractors using combined attribute descriptions.}
    \label{fig:same_img_diff_queries}
\end{figure*}

\begin{figure*}
    \centering
    \includegraphics[width=1\linewidth]{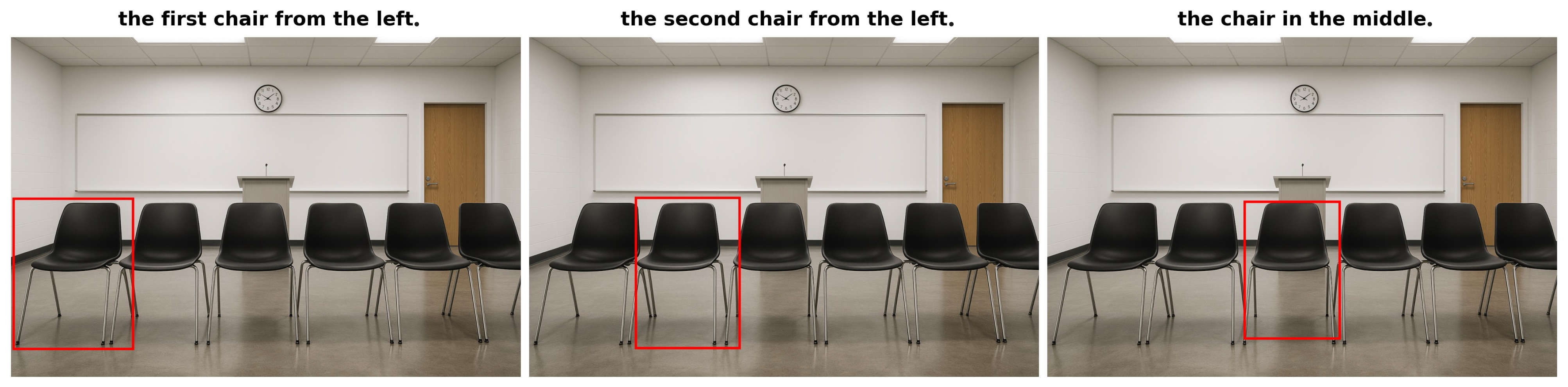}
    \caption{Ordinal reasoning with identical objects. Left: "\textit{the first chair from the left}", Center: "\textit{the second chair from the left}", Right: "\textit{the chair in the middle}". The model successfully performs positional counting to distinguish visually identical chairs based solely on spatial ordering descriptors.}
    \label{fig:ordinal_reasoning}
\end{figure*}

\begin{figure*}
    \centering
    \includegraphics[width=1\linewidth]{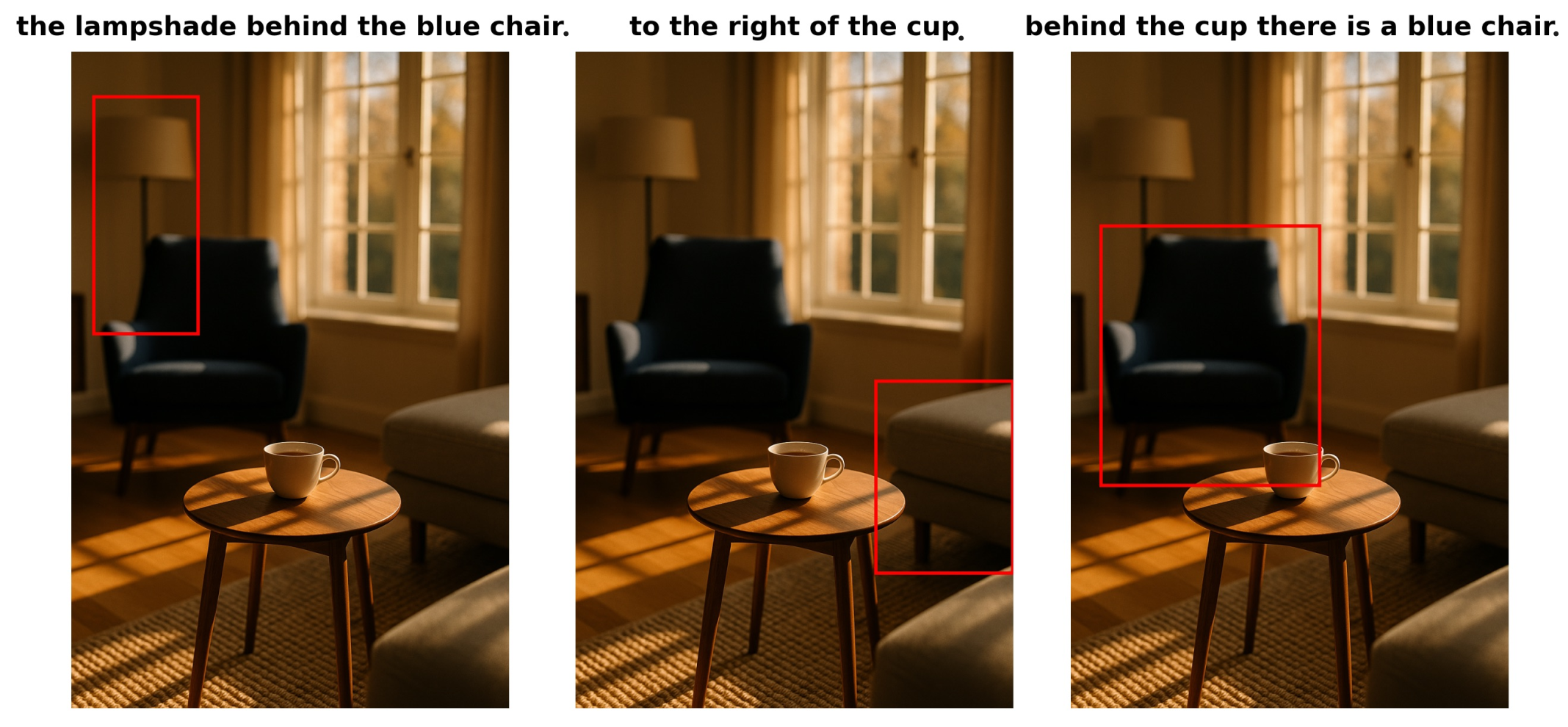}
    \caption{Illustration for multi-hop spatial relationship reasoning. Left: "\textit{the lampshade behind the blue chair}", Center: "\textit{the bench to the right of the cup}", Right: "\textit{behind the cup there is a blue chair}".}
    \label{fig:multihop_spatial}
\end{figure*}

\begin{figure*}
    \centering
    \includegraphics[width=1\linewidth]{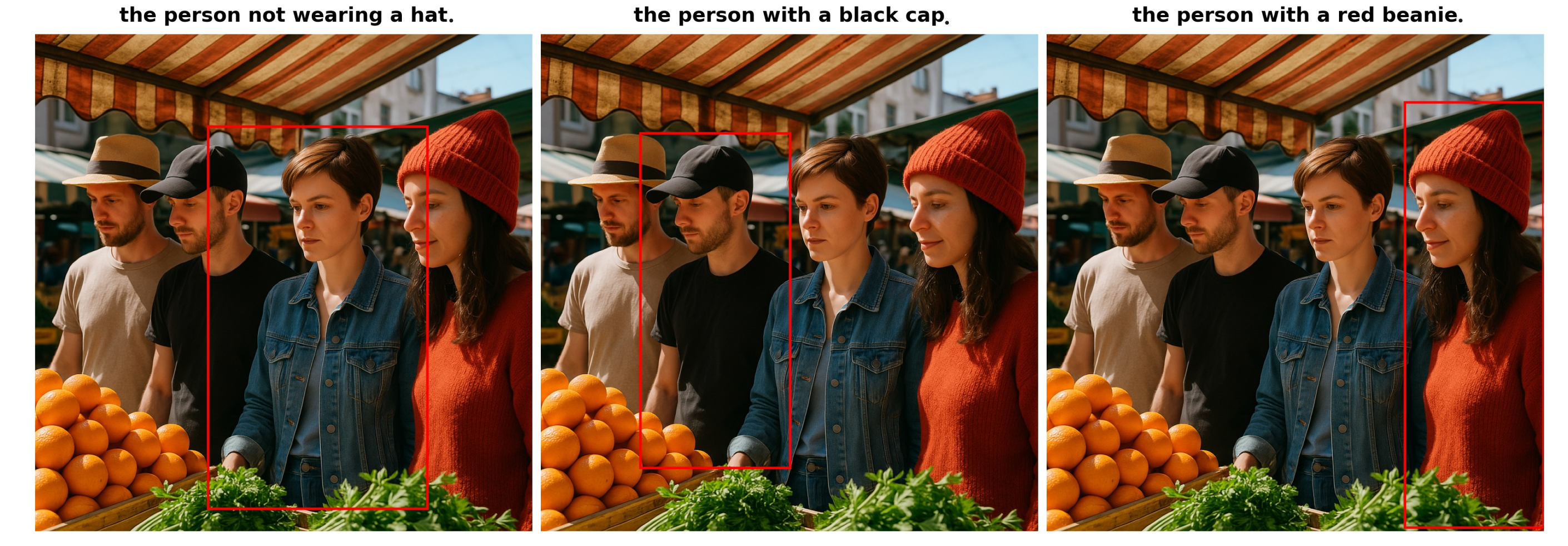}
    \caption{Illustration for negation handling in attribute-based referring expressions. Left: "\textit{the person not wearing a hat}", correctly identifies absence of attribute. Center: "\textit{the person with a black cap}", positive attribute matching. Right: "\textit{the person with a red beanie}", color-specific accessory identification.}
    \label{fig:negation_handling}
\end{figure*}

\begin{figure*}
    \centering
    \includegraphics[width=1\linewidth]{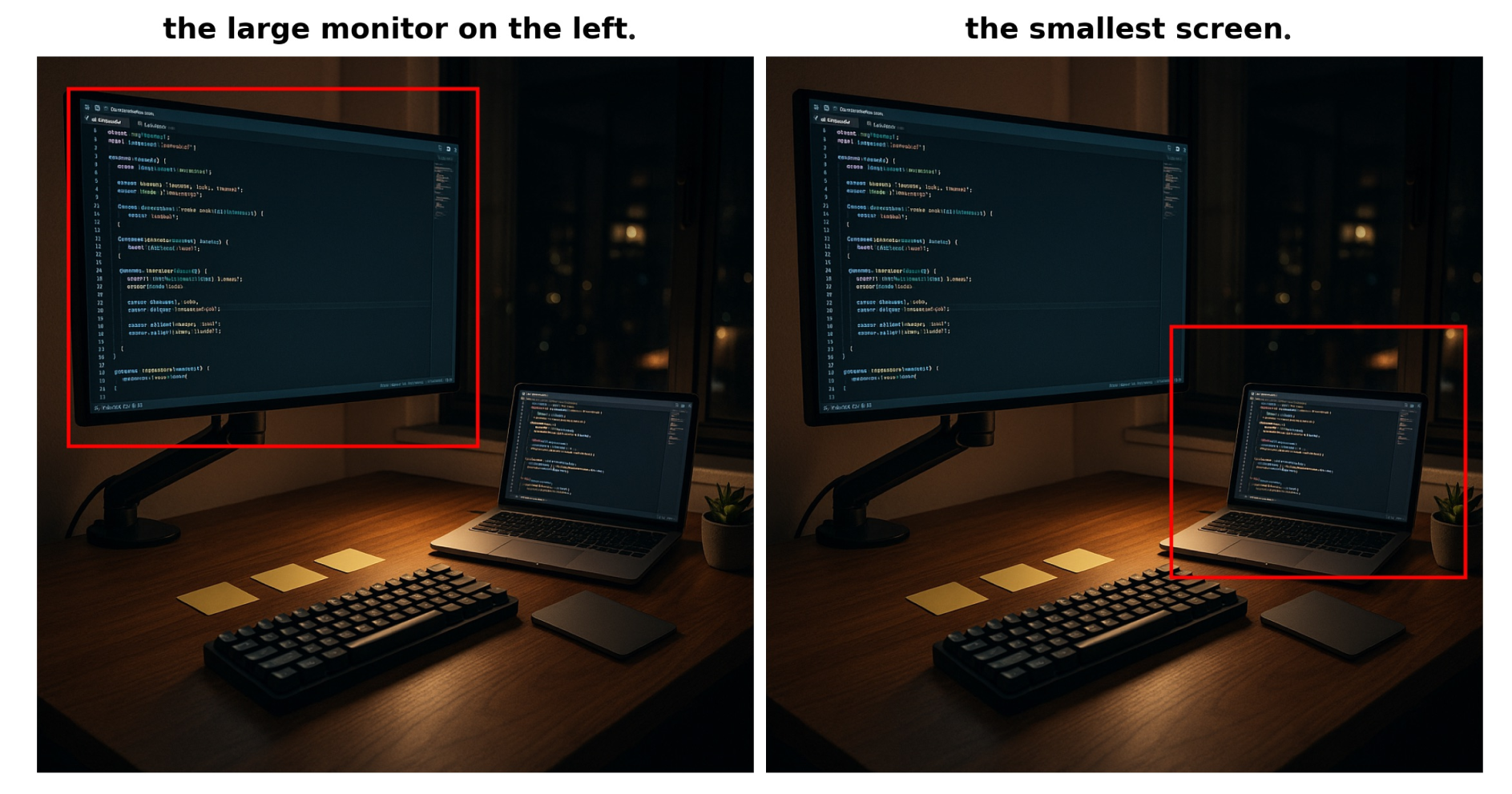}
    \caption{Illustration for relative size discrimination within the same object category. Left: "\textit{the large monitor on the left}", targeting the larger desktop display. Right: "\textit{the smallest screen}", targeting the laptop display.}
    \label{fig:size_discrimination}
\end{figure*}

\begin{figure*}
    \centering
    \includegraphics[width=1\linewidth]{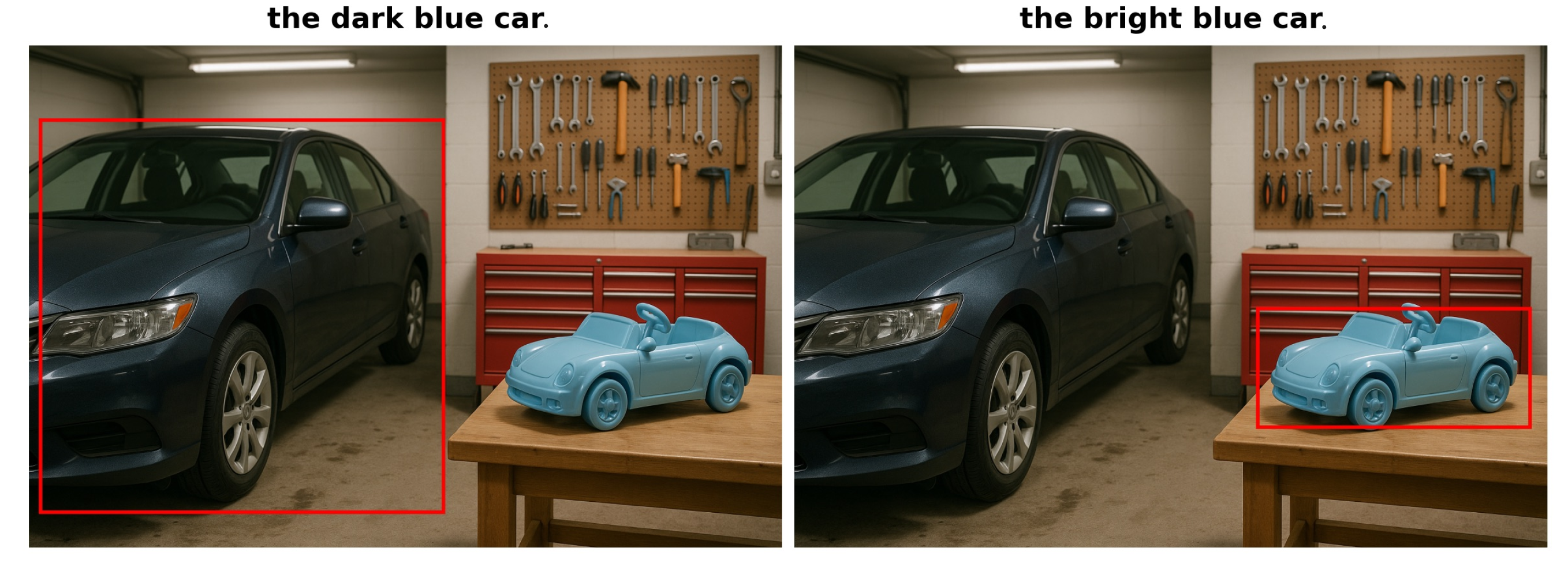}
    \caption{Illustration for fine-grained color discrimination with scale variation. Left: "\textit{the dark blue car}" targeting the full-size vehicle with darker blue paint. Right: "\textit{the bright blue car}" targeting the toy car with lighter blue color.}
    \label{fig:color_discrimination}
\end{figure*}

\begin{figure*}
    \centering
    \includegraphics[width=1\linewidth]{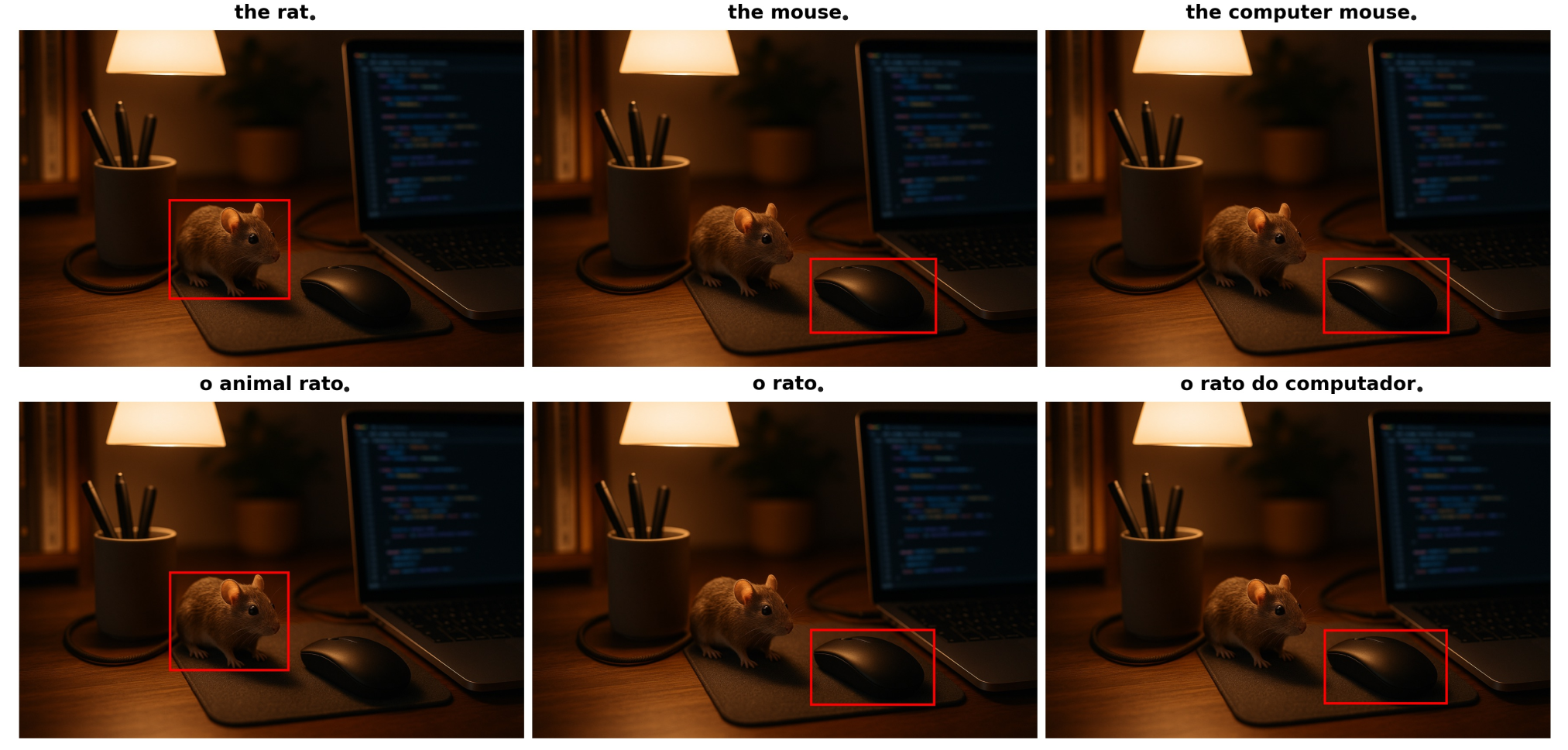}
    \caption{Semantic disambiguation through visual grounding in English and Portuguese. Top row: English "\textit{the rat}" (animal), "\textit{the mouse}" (computer peripheral), and "\textit{the computer mouse}" with explicit category modifier. Bottom row: Portuguese "\textit{o animal rato}" (the animal rat) with category specifier "\textit{animal}", "\textit{o rato}" requiring visual context for disambiguation, and "\textit{o rato do computador}" (the computer mouse/rat) with explicit qualifier "\textit{do computador}". The word "\textit{rato}" in Portuguese inherently references both the biological animal and the computer device, necessitating explicit modifiers for unambiguous reference.}
    \label{fig:semantic_disambiguation}
\end{figure*}

Compositional reasoning capabilities are illustrated through ordinal counting (Figure \ref{fig:ordinal_reasoning}), multi-hop spatial relationships (Figure \ref{fig:multihop_spatial}), negation handling (Figure \ref{fig:negation_handling}), and fine-grained discrimination, including relative size (Figure \ref{fig:size_discrimination}), color differentiation (Figure \ref{fig:color_discrimination}), and semantic disambiguation (Figure \ref{fig:semantic_disambiguation}). In turn, Figure \ref{fig:same_img_diff_queries} demonstrates the model's ability to disambiguate multiple objects within the same office scene through combined attribute descriptions: identifying an earphone by material and color properties, a cup by color and text pattern, and a glass container by material characteristics. These examples collectively validate that the architecture achieves language-agnostic spatial reasoning and complex compositional understanding beyond simple attribute matching.

\begin{figure*}
    \centering
    \includegraphics[width=1\linewidth]{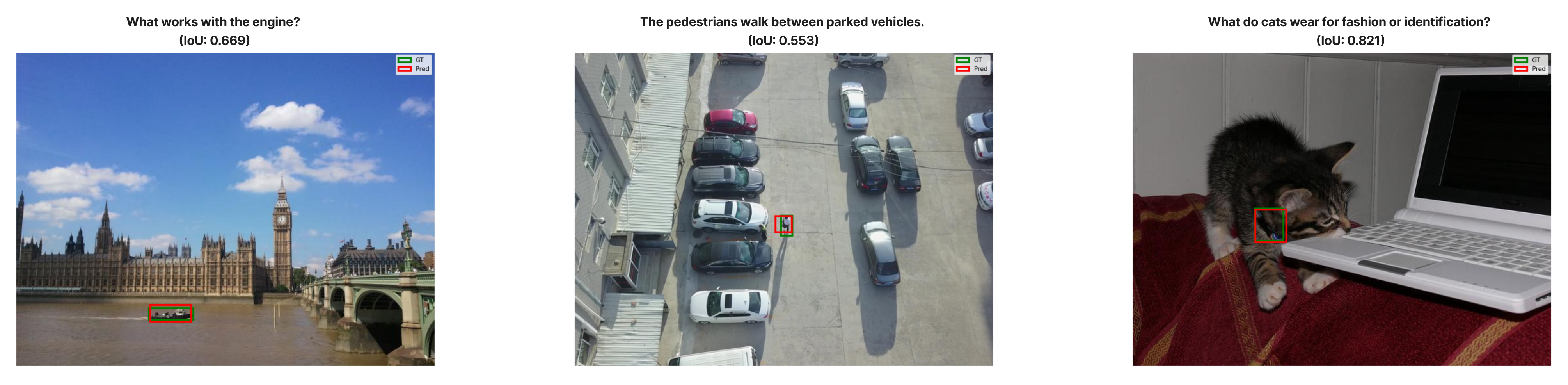}
    \caption{Small object localization examples spanning diverse visual domains. Left: "\textit{What works with the engine?}" targeting a boat on the Thames River (IoU: 0.669); Center: "\textit{The pedestrians walk between parked vehicles}" identifying a small human figure in an aerial parking lot view (IoU: 0.553); Right: "\textit{What do cats wear for fashion or identification?}" localizing a collar on a cat near a laptop (IoU: 0.821). These examples illustrate that localization difficulty correlates with both absolute object size and contextual clutter rather than object category alone.}
    \label{fig:smallexamples}
\end{figure*}

\begin{figure*}
    \centering
    \includegraphics[width=1\linewidth]{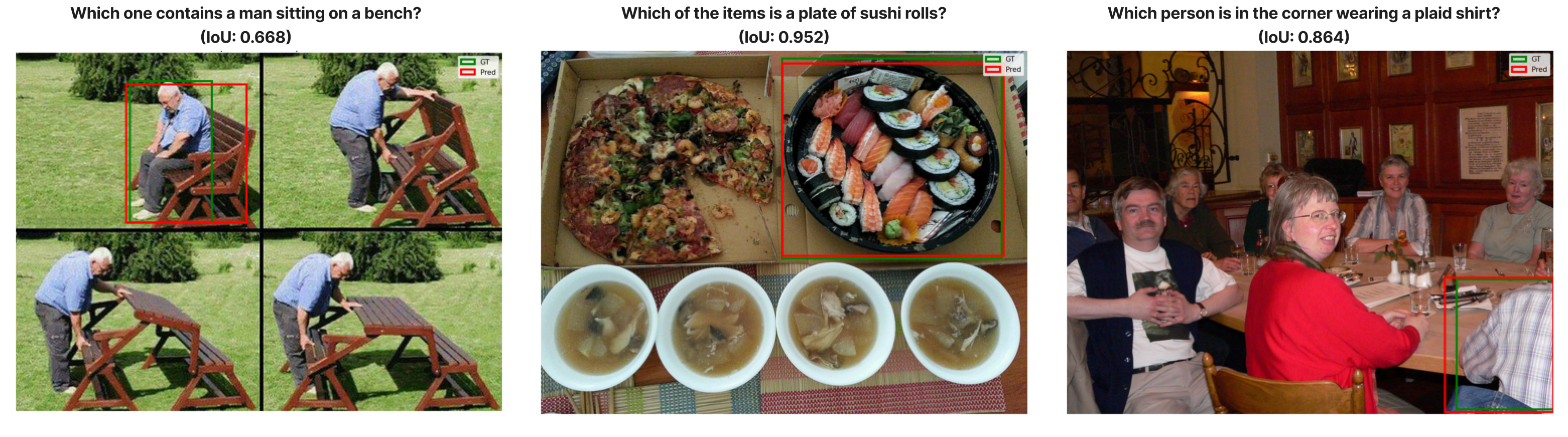}
    \caption{Illustration for model predictions across challenging scenarios: multiple-choice localization with similar objects (left), fine-grained food item identification (center), and person identification with specific attribute descriptions (right). Green boxes indicate the ground truth, while red boxes show predictions.}
    \label{fig:refexpexamples1}
\end{figure*}

Small object localization across domains is presented in Figures \ref{fig:smallexamples} and \ref{fig:refexpexamples1}, illustrating that difficulty correlates with both object size and contextual clutter. 

\begin{figure*}
    \centering
    \includegraphics[width=1\linewidth]{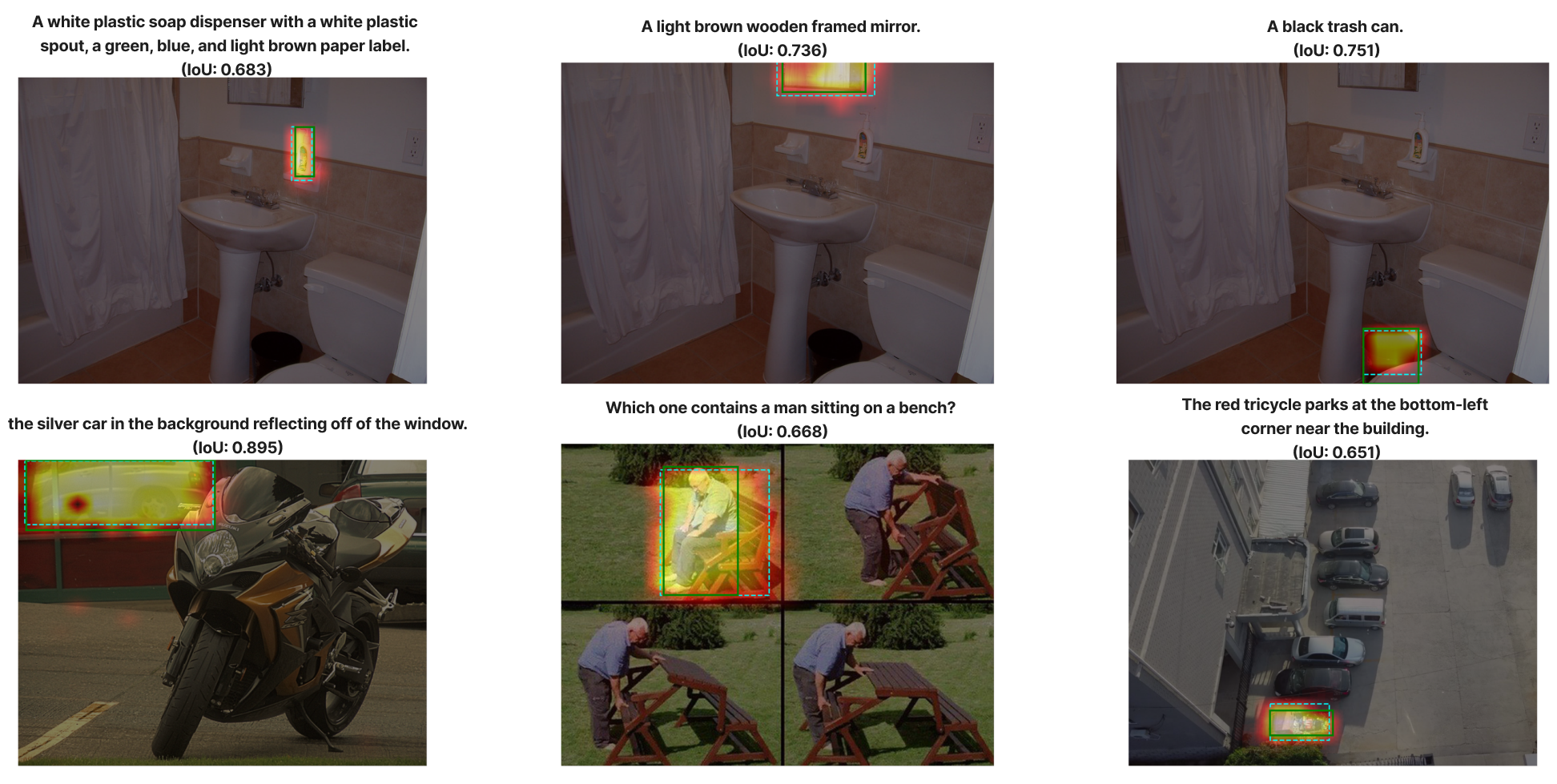}
    \caption{Attention distribution visualizations for five diverse referring expressions. Top row: "\textit{A black trash can}" (IoU: 0.751), "\textit{A light brown wooden framed mirror}" (IoU: 0.736), "\textit{Which one contains a man sitting on a bench?}" (IoU: 0.668). Bottom row: "\textit{The red tricycle parks at the bottom-left corner near the building}" (IoU: 0.651), "\textit{The black cars park along the left side of the parking area}" (IoU: 0.755), "\textit{A white plastic soap dispenser}" (IoU: 0.683).}
    \label{fig:attention_visualization}
\end{figure*}

\begin{figure*}
    \centering
    \includegraphics[width=0.9\linewidth]{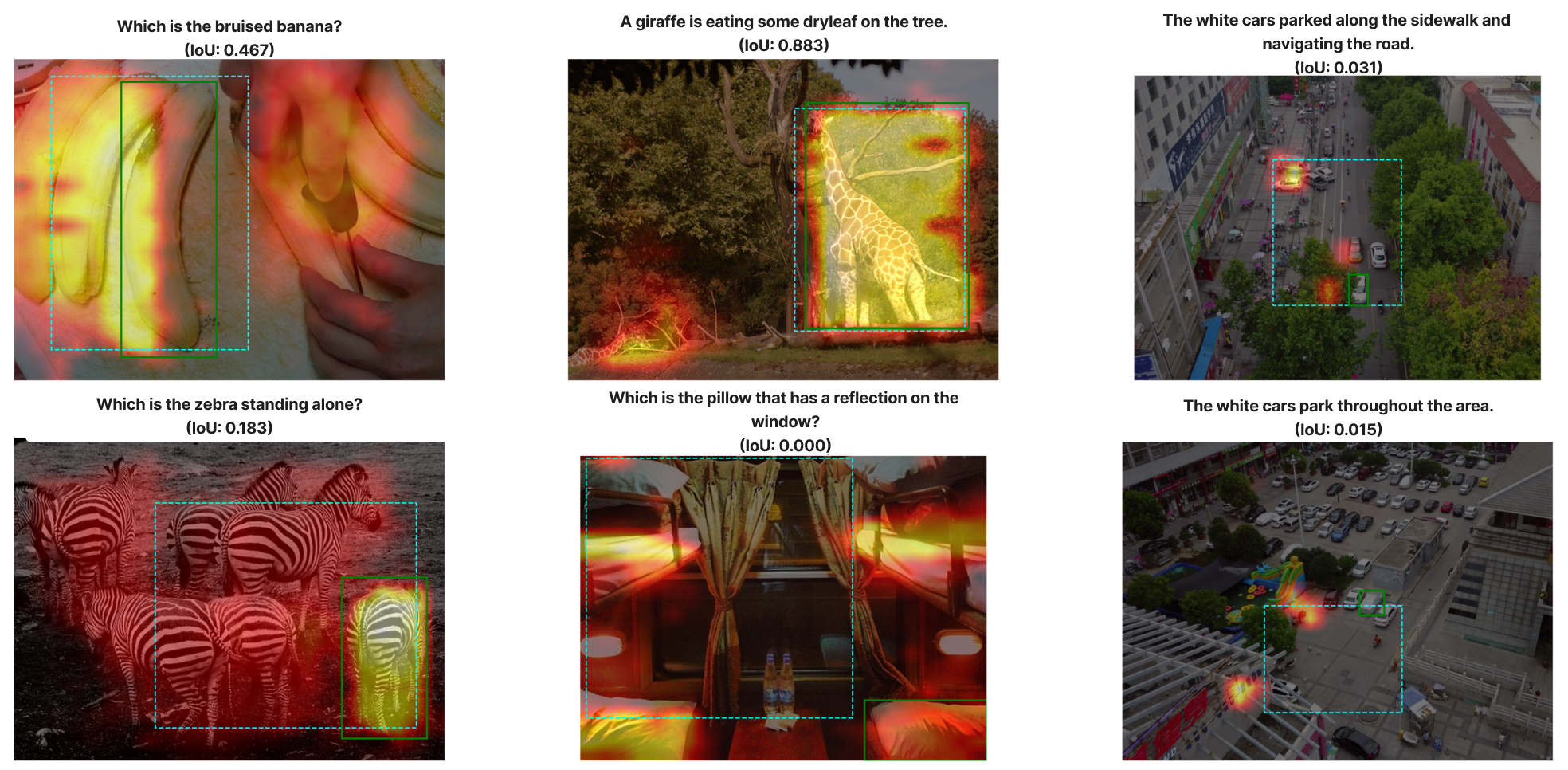}
    \caption{Failure cases illustrating systematic challenges. Top row: "\textit{Which is the bruised banana?}", "\textit{a giraffe is eating some dryleaf on the tree}", "\textit{The white cars parked along the sidewalk and navigating the road.}" Bottom row: "\textit{Which is the zebra standing alone?}", "\textit{Which is pillow that has a reflection on the window?}", "\textit{The white cars park throughout the area.}".}
    \label{fig:multi_attention_patches}
\end{figure*}

\begin{figure*}
    \centering
    \includegraphics[width=1\linewidth]{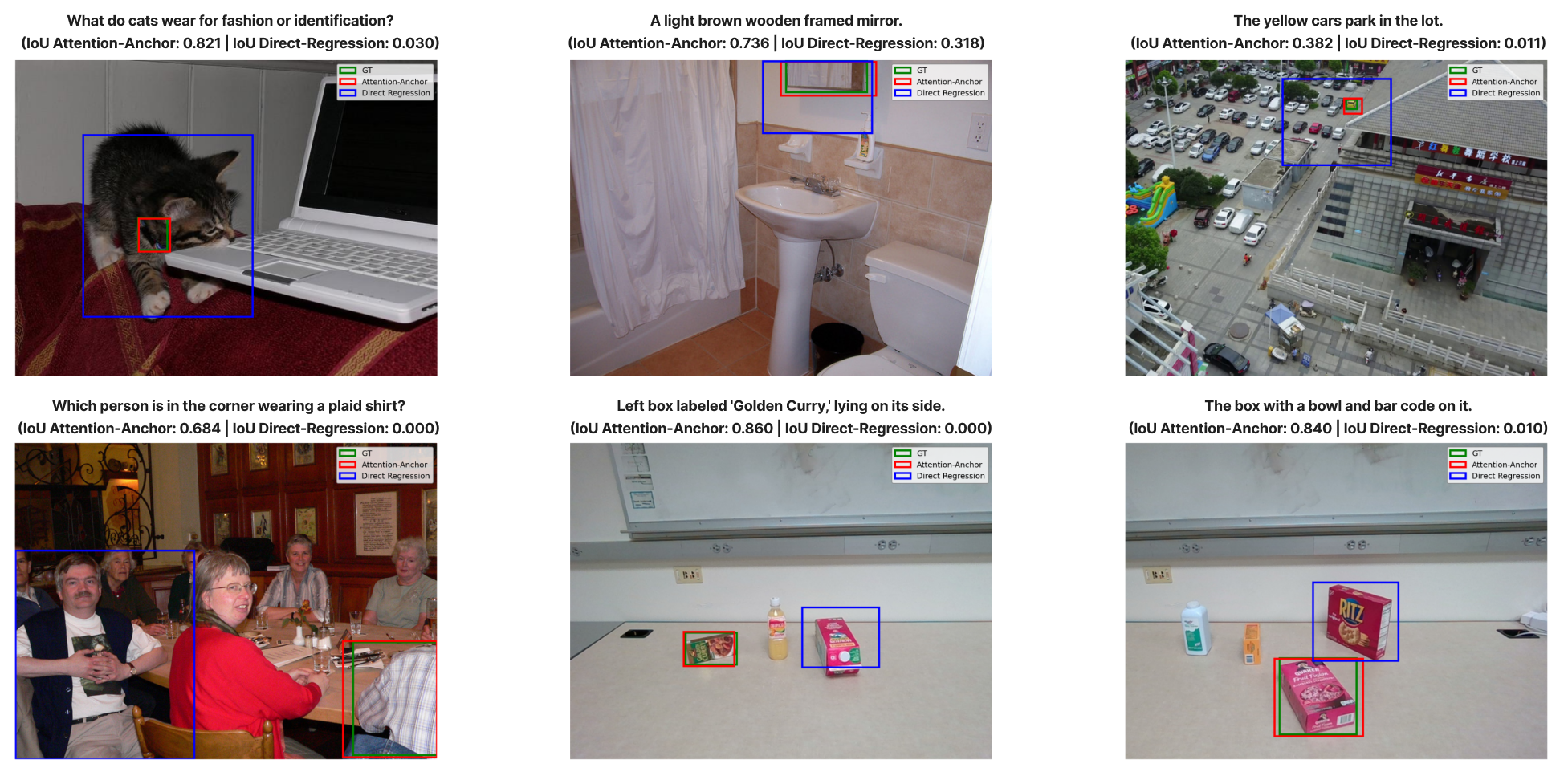}
    \caption{Qualitative comparison between attention-anchored (red) and direct regression (blue) predictions across six scenarios. Green boxes indicate ground truth. Top row: Small object localization where attention-anchored succeeds while direct regression fails. Bottom row: Cases where attention-anchored maintains strong performance while direct regression exhibits complete failures.}
    \label{fig:ablation_qualitative}
\end{figure*}

In turn, Figure \ref{fig:attention_visualization} reveals how attention distributions guide anchor generation, demonstrating interpretable spatial patterns aligned with human intuitions. In complement to this, a qualitative ablation study (Figure \ref{fig:ablation_qualitative}) shows that attention-anchored prediction consistently outperforms direct regression, particularly for small objects and cluttered scenes, validating the architectural decomposition into coarse attention-based anchors followed by learned refinement. Finally, Figure \ref{fig:multi_attention_patches} illustrates failure cases where attention spreads across multiple objects, particularly with plural references, suggesting that explicit singular-plural modeling and improved distractor suppression represent important directions for future work.

\end{document}